\documentclass[10pt,journal,compsoc]{IEEEtran}
\ifCLASSOPTIONcompsoc
  \usepackage[nocompress]{cite}
\else
  \usepackage{cite}
\fi
\ifCLASSINFOpdf
\else
\fi
\usepackage{xcolor}
\usepackage{graphicx}
\usepackage{amsmath}
\usepackage{amssymb}
\usepackage{booktabs}
\usepackage{bbding}
\usepackage{pifont}
\usepackage{algorithm}
\usepackage{algorithmic}
\usepackage{multirow}
\usepackage{booktabs}
\usepackage{enumitem}
\usepackage{makecell}
\usepackage{amssymb}
\usepackage{amsmath}
\usepackage{arydshln}
\usepackage{array}
\usepackage{colortbl}
\DeclareMathOperator*{\argmax}{argmax}

\newcommand{\OK}[1]{\textcolor{black}{#1}} %black

\newcommand{\blue}[1]{\textcolor{blue}{#1}} %blue

\newcommand{\ie}[1]{\textsl{i.e.,}} 
\newcommand{\eg}[1]{\textsl{e.g.,}} 
\newcommand{\etal}[1]{\textsl{et al.}} 

\begin{document}
%
% paper title
% Titles are generally capitalized except for words such as a, an, and, as,
% at, but, by, for, in, nor, of, on, or, the, to and up, which are usually
% not capitalized unless they are the first or last word of the title.
% Linebreaks \\ can be used within to get better formatting as desired.
% Do not put math or special symbols in the title.
\title{ActionHub: A Large-scale Action Video Description Dataset for Zero-shot \\ Action Recognition}

%
%
% author names and IEEE memberships
% note positions of commas and nonbreaking spaces ( ~ ) LaTeX will not break
% a structure at a ~ so this keeps an author's name from being broken across
% two lines.
% use \thanks{} to gain access to the first footnote area
% a separate \thanks must be used for each paragraph as LaTeX2e's \thanks
% was not built to handle multiple paragraphs
%
%
%\IEEEcompsocitemizethanks is a special \thanks that produces the bulleted
% lists the Computer Society journals use for "first footnote" author
% affiliations. Use \IEEEcompsocthanksitem which works much like \item
% for each affiliation group. When not in compsoc mode,
% \IEEEcompsocitemizethanks becomes like \thanks and
% \IEEEcompsocthanksitem becomes a line break with idention. This
% facilitates dual compilation, although admittedly the differences in the
% desired content of \author between the different types of papers makes a
% one-size-fits-all approach a daunting prospect. For instance, compsoc 
% journal papers have the author affiliations above the "Manuscript
% received ..."  text while in non-compsoc journals this is reversed. Sigh.

\author{Jiaming Zhou, Junwei Liang, Kun-Yu Lin, Jinrui Yang, Wei-Shi Zheng$^\dagger$% <-this % stops a space
\thanks{$\dagger$Corresponding author: Wei-Shi Zheng}
\IEEEcompsocitemizethanks{
\IEEEcompsocthanksitem Jiaming Zhou is with School of Computer Science and Engineering, Sun Yat-sen University, Guangzhou, China. This work is partially done when Jiaming is with Hong Kong University of Science and Technology (Guangzhou), China. (e-mail: jia\_ming\_zhou@outlook.com)
\IEEEcompsocthanksitem Junwei Liang is with AI Thrust, Hong Kong University of Science and Technology (Guangzhou) and also affiliated with CSE, Hong Kong University of Science and Technology. (e-mail: junweiliang@hkust-gz.edu.cn).
\IEEEcompsocthanksitem Kun-Yu Lin is with School of Computer Science and Engineering, Sun Yat-sen University, Guangzhou, China. (e-mail: linky5@mail2.sysu.edu.cn)
\IEEEcompsocthanksitem Jinrui Yang is with Computer Science and Engineering Department, University of California, Santa Cruz. (e-mail: jyang347@ucsc.edu)
\IEEEcompsocthanksitem Wei-Shi Zheng is with the School of Computer Science and Engineering and the Key Laboratory of Machine Intelligence and Advanced Computing, Ministry of Education, Sun Yat-sen University, Guangzhou 510275, China, and also with the Peng Cheng Laboratory, Shenzhen 518005, China. (e-mail: wszheng@ieee.org)
}% <-this % stops an unwanted space
}

% The paper headers
\markboth{Submission to IEEE Transactions on XXX}%
{}

\IEEEtitleabstractindextext{%
\begin{abstract}
Zero-shot action recognition (ZSAR) aims to learn an alignment model between videos and class descriptions of seen actions that is transferable to unseen actions. 
There is a rich diversity in video content, including complex scenes, dynamic human motions, etc. 
The text queries (class descriptions) used in existing ZSAR works, however, are often short action names that fail to capture the rich semantics in the videos, leading to misalignment. With the intuition that video content descriptions (\eg, video captions) can provide rich contextual information of visual concepts in videos, which helps model understand human actions from text modality, we propose to utilize human annotated video descriptions to enrich the semantics of the class descriptions of each action. However, all existing action video description datasets are limited in terms of the number of actions, the diversity of actions, the semantics of video descriptions, etc. To this end, we collect a large-scale action video descriptions dataset named \textbf{ActionHub}, which covers a total of 1,211 common actions and provides 3.6 million action video descriptions. With the proposed ActionHub dataset, we find that the semantics of human actions can be better captured from the textual modality, such that the cross-modality diversity gap between videos and texts in ZSAR is alleviated, and a transferable alignment is learned for recognizing unseen actions. To achieve this, we further propose a novel \textbf{C}ross-m\textbf{o}dality and \textbf{C}ross-acti\textbf{o}n Modeling (\textbf{CoCo}) framework for ZSAR, which consists of a Dual Cross-modality Alignment module and a Cross-action Invariance Mining module. Specifically, the Dual Cross-modality Alignment module utilizes both action labels and video descriptions from ActionHub to obtain rich class semantic features for feature alignment. The Cross-action Invariance Mining module exploits a cycle-reconstruction process between the class semantic feature spaces of seen actions and unseen actions, aiming to guide the model to learn cross-action invariant representations. Extensive experimental results demonstrate that our CoCo framework significantly outperforms the state-of-the-art on three popular ZSAR benchmarks (\ie, Kinetics-ZSAR, UCF101 and HMDB51) under two different learning protocols in ZSAR, proving the efficacy of our proposed dataset and method. We will release our code, models, and the proposed ActionHub dataset.

\end{abstract}

% Note that keywords are not normally used for peerreview papers.
\begin{IEEEkeywords}
Action Recognition, Zero-shot Action Recognition, Action Video Description Dataset. 
\end{IEEEkeywords}}

% make the title area
\maketitle

% To allow for easy dual compilation without having to reenter the
% abstract/keywords data, the \IEEEtitleabstractindextext text will
% not be used in maketitle, but will appear (i.e., to be "transported")
% here as \IEEEdisplaynontitleabstractindextext when the compsoc 
% or transmag modes are not selected <OR> if conference mode is selected 
% - because all conference papers position the abstract like regular
% papers do.
\IEEEdisplaynontitleabstractindextext
% \IEEEdisplaynontitleabstractindextext has no effect when using
% compsoc or transmag under a non-conference mode.

% For peer review papers, you can put extra information on the cover
% page as needed:
% \ifCLASSOPTIONpeerreview
% \begin{center} \bfseries EDICS Category: 3-BBND \end{center}
% \fi
%
% For peerreview papers, this IEEEtran command inserts a page break and
% creates the second title. It will be ignored for other modes.
\IEEEpeerreviewmaketitle

\begin{figure*}[!th]
\centering
\includegraphics[width=0.85\textwidth]{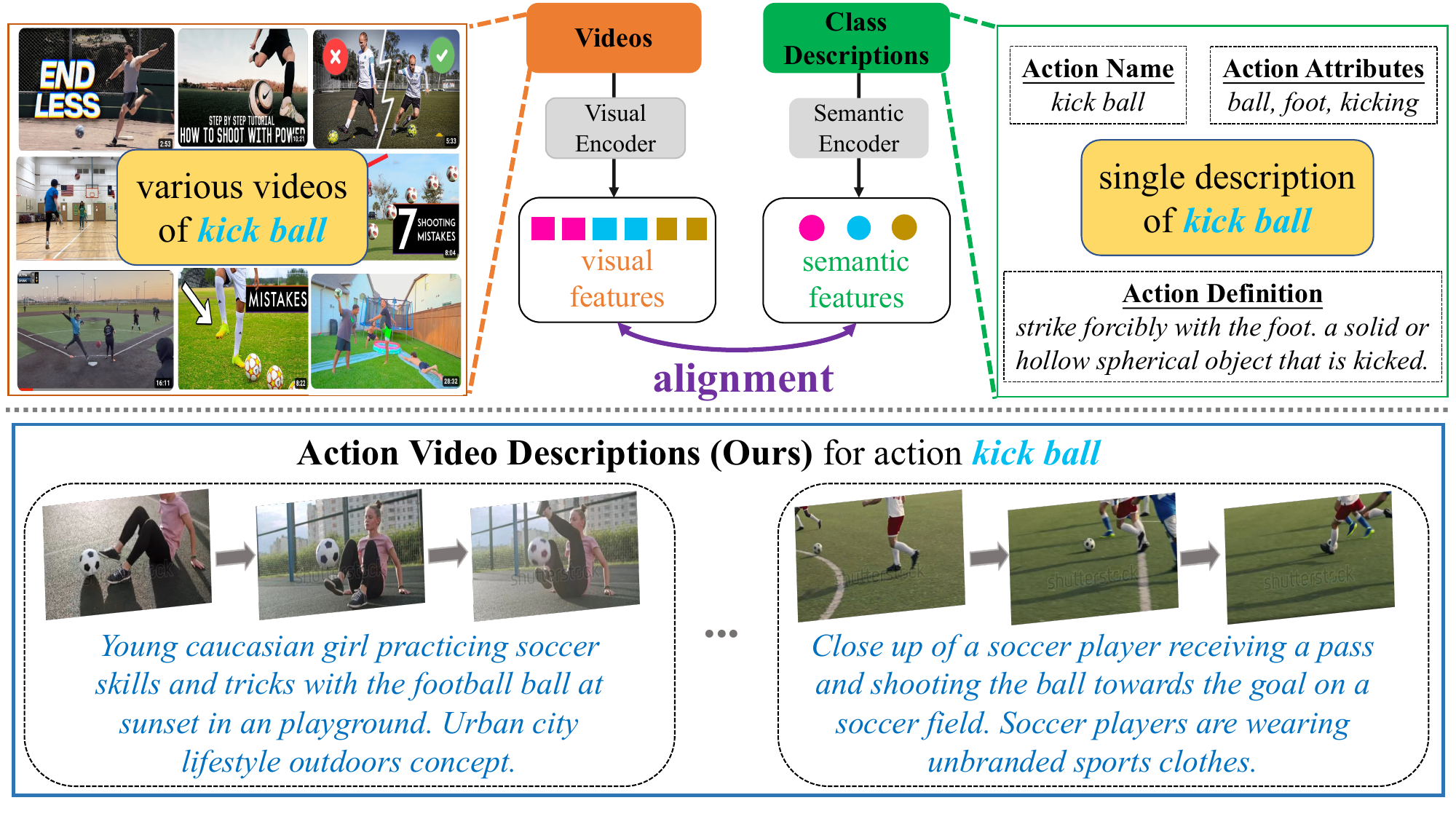}
\caption{The top of the figure shows a common framework for zero-shot action recognition, which maps the videos and class descriptions of actions into the corresponding feature spaces, and then learns an alignment between these two feature spaces. The class descriptions used in existing works are generally action names, action attributes, and action definitions. The semantics of these class-level descriptions are limited when matched with the rich semantics of those visual concepts in videos (\eg, action performer, objects, scenes), leading to the cross-modality diversity gap between videos and texts. The bottom of the figure shows two instances of action video descriptions for the action "kick ball", which provide abundant textual information correlated with the visual concepts in the videos of the action, thus the cross-modality diversity gap can be effectively alleviated.}
\label{motivation}
\end{figure*}

\IEEEraisesectionheading{\section{Introduction}\label{sec:intro}}
\IEEEPARstart{V}{ideo} action recognition aims to recognize the actions performed by humans in videos. In recent years, many advanced video backbones~\cite{wang2016temporal, wang2018non, lin2019tsm, feichtenhofer2019slowfast, Zhou_2021_CVPR, bertasius2021space, fan2021multiscale, liu2021video} have been proposed, which have greatly promoted the development of this field. However, the actions that need to be recognized in real-world scenarios are constantly emerging. 
To recognize new actions, existing supervised learning framework needs to annotate a large number of videos and retrain a supervised action recognition model, which is time-consuming and resource-intensive. 
Therefore, zero-shot action recognition (ZSAR) is proposed to train a transferable model using the data of seen actions, so that it can generalize to unseen actions for action recognition without annotating new training data. As shown in the top of Figure~\ref{motivation}, a common framework for ZSAR is to first map the videos and class descriptions (\eg, action names) of seen actions into the visual space and semantic space, respectively. Then an alignment is learned to embed the visual and semantic features into a joint space. 
Ideally, this framework can generalize to unseen actions given new action names.

Figure~\ref{motivation} shows that the class descriptions used to linguistically define human actions in existing ZSAR works include action names, action attributes, and action definitions~\cite{chen2021elaborative}, etc. These class descriptions are annotated to define actions using either a few words or sentences. However, human action videos contain various visual concepts, \eg, scenes, objects, and human motion, which make the video modality rich in semantic diversity. Existing class descriptions lack the textual context information to describe the corresponding visual concepts in videos. Such class-level annotations are ambiguous in semantics to match various human action instances in videos, which causes the \textbf{\textit{cross-modality diversity gap}} between the video and the text modalities, making the alignments learned by existing ZSAR works still far from satisfactory. 
Intuitively, the descriptions of video content (\eg, video captions) contain a lot of textual context information about videos, which can be modeled to correlate with the semantics of visual concepts in videos, so that the model understands human actions more easily from the text modality. The bottom of Figure~\ref{motivation} shows two instances of action video descriptions of the action \textit{kicking ball}, which reveals the context of various visual concepts that may appear in the action from the text modality (\eg, the descriptions of how the ball, players, and kicking motion evolve in videos), such that the semantics of the action can be effectively modeled. By providing such rich action video descriptions for each action, the cross-modality diversity gap between videos and class descriptions can be alleviated, thus the further learned alignment can better generalize to the unseen actions.

To achieve this, it is essential to have a video dataset that not only covers large-scale human actions, but also provides textual descriptions to describe the actions performed in videos. Unfortunately, the current video understanding community fails to provide such a large-scale action video description dataset. The existing large-scale video caption datasets (\eg, VideoCC3M~\cite{nagrani2022learning}) are generally collected from open domains (\ie, encompass a broad spectrum of visuals and topics) instead of the specific action domain, which cannot provide the textual data that describes human actions in videos. And the video action datasets collected from the action domain either lack descriptions of video content (\eg, Kinetics700~\cite{carreira2019short}), or fail to meet the requirement of covering large-scale human actions (\eg, ActivityNet-Cap~\cite{krishna2017dense} only has 200 actions).

To this end, we collect a large-scale action video description dataset named \textbf{ActionHub}, aiming to solve the cross-modality diversity gap between videos and texts in zero-shot action recognition. The ActionHub dataset contains over 3 million video descriptions for 1211 human actions. Specifically, for each action, we use its action name as a search query on video websites, and collect the video descriptions of the returned videos (provided by the websites) without the need for additional annotations, making the ActionHub a low-cost and highly scalable dataset.

\begin{figure*}[!th]
\centering
\includegraphics[width=0.95\textwidth]{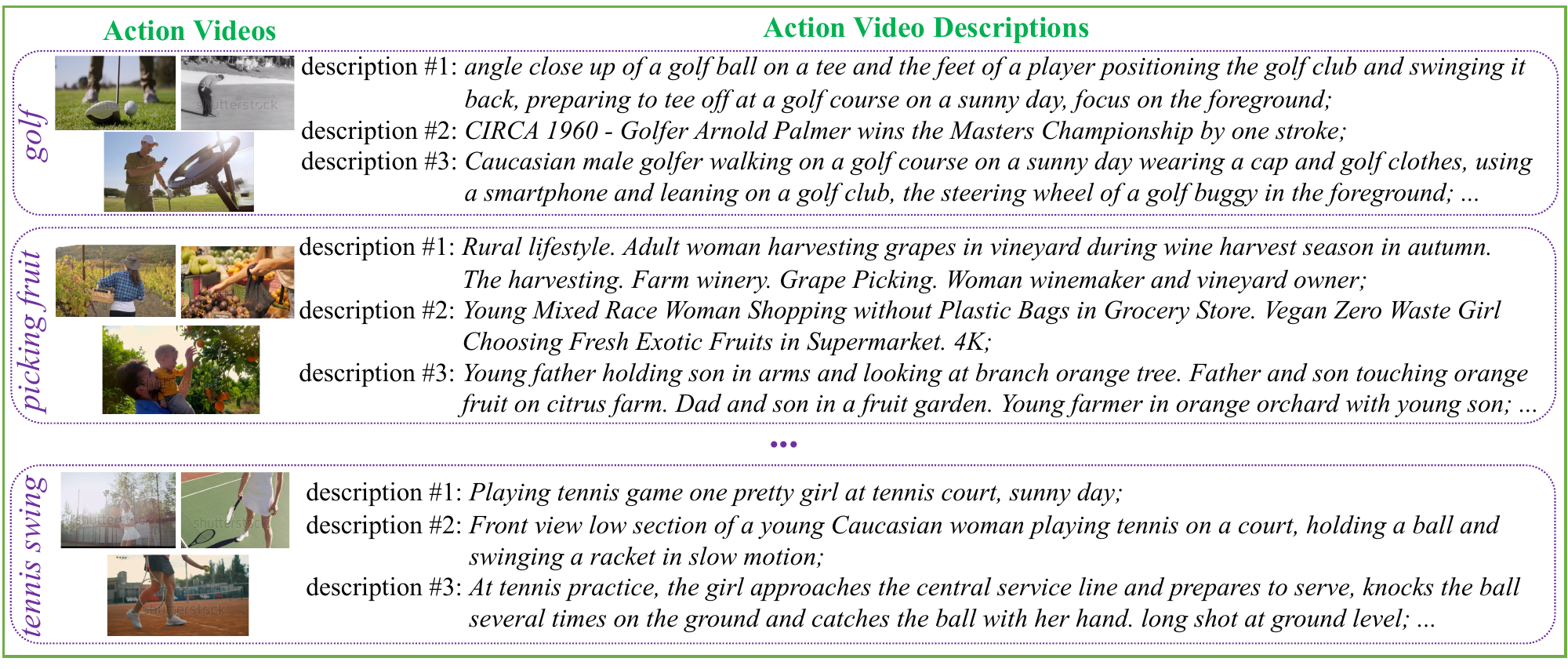}
\caption{Three actions, \ie, "golf", "picking fruit", and "tennis swing" from the proposed ActionHub dataset. For each action, we show three instances of action video descriptions, which provide rich contextual information about the visual concepts of videos of the action. Such rich textual semantics of actions in our proposed ActionHub dataset help ZSAR models better understand human actions.}
\label{examples_for_ActionHub}
\end{figure*}

The proposed ActionHub dataset enables the learning of an effective alignment between video and text data of actions, by providing the action video descriptions to mitigate the cross-modality diversity gap. Notably, our ActionHub dataset is the largest action video dataset with action content descriptions to date. The proposed ActionHub dataset covers large-scale common human actions in real scenarios, by building action queries using action names from seven existing video action datasets. 
Compared to existing action video description datasets, our ActionHub dataset shows its superiority in the following aspects: it covers the largest number of actions to date, collects diverse actions from multiple domains, has high scalability for collecting more actions, and is equipped with well-matched descriptions with rich semantics (see Sec.~\ref{motivation_for_actionhub} for details).
The release of the ActionHub dataset can effectively enrich the semantic diversity of text modality of actions, which is the prerequisite for learning a generalizable alignment for human actions. Figure~\ref{examples_for_ActionHub} shows some examples from the collected ActionHub dataset. 
By leveraging the resources and insights provided by ActionHub, we expect to push the boundaries of what is currently achievable in the realm of zero-shot action recognition, and pave the way for innovative solutions in the future.

Based on the proposed ActionHub dataset, we propose a \textbf{C}ross-m\textbf{o}dality and \textbf{C}ross-acti\textbf{o}n Modeling (\textbf{CoCo}) framework for zero-shot action recognition. The CoCo framework consists of a Dual Cross-modality Alignment module and a Cross-action Invariance Mining module. 
The Dual Cross-modality Alignment module utilizes both existing action definitions and video descriptions from ActionHub to enrich the diversity of action class descriptions, thus better class semantic features can be obtained for feature alignment.
Moreover, there are various common semantic concepts (\eg, objects, sub-actions, and human-object interactions) across seen and unseen actions. And the semantics of each action class is composed of these cross-action invariant semantic concepts. Thus learning the invariant and discriminative representations of these semantic concepts can further boost the alignment for each action class. Previous works are limited to utilizing manual-defined elements~\cite{mettes2021object, gao2019know} (\eg, objects, attributes, nouns and verbs) or information interaction between different actions~\cite{ghosh2021learning, 9173643}, which lack explicit feature constraints.% and do not address the seen-unseen bias~\cite{brattoli2020rethinking,chen2021elaborative}. 
To this end, our CoCo framework introduces the Cross-action Invariance Mining module, which utilizes the class semantic feature space of unseen actions to cycle-reconstruct that of seen actions. By constraining the semantic consistency of the reconstructed class semantic features, our model learns discriminative features that are invariant across seen and unseen actions.
By reducing the cross-modality diversity gap and mining the cross-action invariance, the proposed CoCo framework effectively learns a transferable alignment between visual and semantic spaces. 

To sum up, the contributions of this work are as follows:
\begin{itemize}[itemsep=0pt, topsep=0pt, parsep=0pt, partopsep=0pt]
    \item A large-scale action video description dataset named \textbf{ActionHub} is proposed, which is the first, and the largest dataset that provides millions of video descriptions to describe thousands of human actions. The proposed ActionHub dataset can help promote the understanding of human actions in the research community.

    \item A Cross-modality and Cross-action (\textbf{CoCo}) framework is proposed, which exploits the proposed ActionHub dataset to reduce the cross-modality diversity gap and mine the cross-action invariance, effectively learning a transferable alignment for zero-shot action recognition.
\end{itemize}
Our CoCo framework achieves state-of-the-art performance on three popular zero-shot action recognition benchmarks (\ie, Kinetics-ZSAR~\cite{chen2021elaborative}, UCF101~\cite{ucf101}, HMDB51~\cite{hmdb51}), under a comprehensive comparison of existing two different learning protocols in zero-shot action recognition.

\section{Related Work}

\subsection{Video Action Recognition}
In supervised action recognition, the research community has advanced with both new datasets and new models. 
The modern action recognition datasets have grown in both the number of classes and the number of videos.
Early benchmarks like UCF-101~\cite{ucf101} and HMDB51~\cite{hmdb51} contained a few thousand videos and action classes of about a hundred.
The modern benchmarks for action recognition include the Kinetics dataset~\cite{kay2017kinetics} and Moments-in-Time~\cite{monfort2019moments}, which contain more categories and more videos (\eg, 400 classes and more than 240K clips in~\cite{kay2017kinetics} and 700 classes in~\cite{carreira2019short}).
However, the number of classes in these datasets is still far from covering all possible actions in different scenarios.
For example, surveillance actions are missing in the two datasets. 
In terms of models, recent works adopted 2D and 3D convolution neural networks~\cite{tran2018closer,lin2019tsm,feichtenhofer2019slowfast,liang2020spatial} and achieved reasonable results. 
These models require a large number of training samples for each action class, even more so for the recent data-hungry vision transformer models~\cite{liu2021video,fan2021multiscale,arnab2021vivit, 10210078, liang2022stargazer,liang2022multi}.
The labor-intensive and time-consuming annotations of massive action videos motivate the community to develop action recognition models with less supervision~\cite{iosifidis2014semi, 9577597, liu2011recognizing, zhou2023adafocus, lin2023diversifying}. 
In this work, we focus on the zero-shot action recognition task.

\begin{figure*}[!t]
\centering
\includegraphics[width=0.9\textwidth]{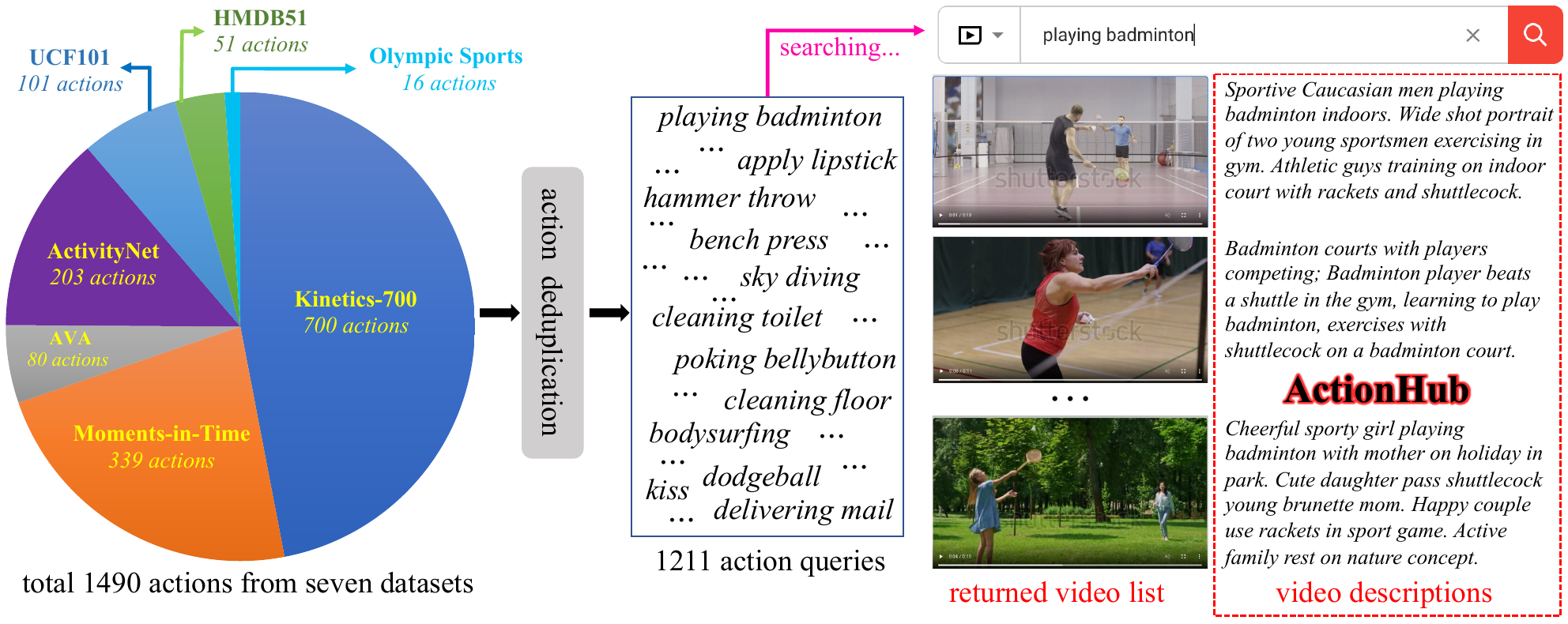}
\caption{The process of collecting the ActionHub dataset from the Internet. To collect a large-scale action video description dataset, we select a total of 1490 actions from seven popular video action datasets (\ie, Kinetics-700~\cite{carreira2019short}, Moments-in-Time~\cite{monfort2019moments}, AVA~\cite{gu2018ava}, ActivityNet~\cite{caba2015activitynet} Olympic Sports~\cite{niebles2010modeling}, HMDB51~\cite{hmdb51}, and UCF101~\cite{ucf101}). After action deduplication, we obtain 1211 action queries. For each action, we use the action name as query to search videos from websites. The descriptions of videos in the returned video list (provided by the websites) are kept as the ActionHub dataset.}
\label{collection_of_ActionHub}
\end{figure*}

\subsection{Zero-shot Action Recognition}

To solve the problem of lack of action data, zero-shot action recognition (ZSAR) task~\cite{liu2011recognizing} is proposed. Existing ZSAR works usually aim to learn an alignment model between the learned visual space of videos and the semantic space of class descriptions. 

\vspace{0.1cm}

\noindent \textbf{Visual and Semantic Feature Learning.} For visual feature learning, early ZSAR works used various hand-crafted features~\cite{liu2011recognizing, kodirov2015unsupervised}. Later, many works~\cite{jain2015objects2action, gan2016concepts, piergiovanni2020learning} extracted visual features with richer semantics from various deep visual networks. The semantic richness of existing class descriptions is weaker than that of videos, which causes the cross-modality diversity gap between videos and text data of human actions. Thus, learning a rich semantic space is more effective for ZSAR. For semantic feature learning, many works~\cite{fu2014transductive, wang2017zero, hahn2019action2vec} utilized the manually-defined action names or attributes as the class descriptions of actions. Semantic hierarchies~\cite{rohrbach2012script} were mined for subjects, verbs, and objects to represent the action labels. ER~\cite{chen2021elaborative} expanded the action classes as sentences (\ie, action definitions), which are more discriminative than class names. However, the class descriptions used in these works still lack the textual context information corresponding to videos, leading to the misalignment of features. In this work, we propose the ActionHub dataset, which provides textual descriptions of visual context in videos, such that the cross-modality diversity gap in ZSAR is effectively alleviated. 

\noindent \textbf{Transferable Alignment Learning.} For both visual and semantic features, existing works attempted to enhance their transferability, thus the learned ZSAR models generalize better on unseen actions. 
TS-GCN~\cite{gao2019know} modeled the semantic relationships among actions and attributes. Some works~\cite{jain2015objects2action, mettes2021object} utilized pre-trained object detectors to determine objects and then computed similarities between objects and actions. Gan \textsl{et al.}~\cite{gan2016learning} treated each category as a domain, aiming to learn cross-category generalizable attribute detectors. ER model~\cite{chen2021elaborative} enforced the model to rehearse video contents with additional semantic knowledge from detected objects to improve the generalization of unseen actions. All these works are limited to utilizing manual-defined elements and lack explicit feature constraints to learn cross-action invariant semantics. In this work, we propose to constrain the semantic consistency of the reconstructed class semantic features, which improves the transferability of the learned features explicitly.

\vspace{0.1cm}

\noindent \textbf{Two Different Learning Protocols in ZSAR.} All the aforementioned works followed an intra-dataset learning protocol in ZSAR, \ie, the seen actions for training and unseen actions for test come from the same video action dataset. Recently, several ZSAR works~\cite{brattoli2020rethinking, pu2022alignment, lin2022cross, kerrigan2021reformulating, Doshi_2023_CVPR} applied a cross-dataset learning protocol in ZSAR, which samples seen actions from a large-scale action dataset (\eg, Kinetics-400~\cite{kay2017kinetics}) for training, and samples unseen actions from other action datasets (\eg, UCF-101~\cite{ucf101}) for testing. This data-hungry learning protocol requires ZSAR models to mine more powerful feature representations. To this end, all these works~\cite{brattoli2020rethinking, pu2022alignment, lin2022cross, kerrigan2021reformulating, Doshi_2023_CVPR} trained their backbone networks end-to-end on the large-scale action dataset, which is time-consuming and resource-intensive. In this work, we mainly follow the more commonly used intra-dataset learning protocol in ZSAR, which does not train the backbone networks but focuses more on the design of feature alignment. Nevertheless, this work still makes a comprehensive comparison under both these two learning protocols, which fully demonstrates the effectiveness of our ActionHub dataset and CoCo framework.

% \subsection{General Zero-shot Learning}
% optional

\subsection{Vision-Language Pretraining for Generalization}
Recently, various vision-language foundation models~\cite{radford2021learning, li2022blip, alayrac2022flamingo, xu2021videoclip, rasheed2023fine, lin2022frozen, wang2023all, Cheng_2023_CVPR, Huang_2023_CVPR} have demonstrated their powerful and generalizable capabilities across different visual tasks. These models require large-scale image-text or video-text data for pretraining. In the domain of action videos, many foundation models have been developed, primarily following two paths. The first path~\cite{xu2021videoclip, rasheed2023fine, lin2022frozen} extends the capabilities of existing image-based foundation models for action understanding by fine-tuning or adapting them using action video datasets. The second path~\cite{wang2023all, Cheng_2023_CVPR, Huang_2023_CVPR} involves training video-based foundation models directly on large-scale video-language data collected from various sources.

All of these models exhibit remarkable results in the zero-shot test. However, they are trained on massive data, and the downstream zero-shot benchmarks are not strictly constructed in the same manner as the traditional zero-shot setting presented in this work. Therefore, this work exclusively focuses on models developed within the traditional zero-shot setting for the purpose of fair comparison.

\subsection{Human Action Understanding with Additional Data}
Most video action understanding works only focused on the frame sequences sampled from the raw video data, which limits the model's ability to understand human actions. Recently, many works have exploited various additional data to facilitate action understanding, which includes visual data (\eg, skeleton~\cite{shi2019skeleton}, depth~\cite{rahmani2017learning}, and point cloud~\cite{cheng2016orthogonal}), audio data~\cite{gao2020listen, damen2020epic}, text data~\cite{wang2022omnivl, xu2021videoclip}, and others (\eg, acceleration~\cite{kwapisz2011activity} and radar~\cite{wang2015understanding}).

Similar to our work, many previous ZSAR works also utilized additional text data to help models recognize unseen human actions in videos. Some ZSAR works used object names~\cite{gao2019know, bretti2021zero} or manual-defined action attribute names~\cite{alexiou2016exploring, gao2019know} to enhance the generalization of the learned feature representations. However, the annotations for such additional text data (\eg, action attributes) are expensive and ambiguous. ER model~\cite{chen2021elaborative} also extended the short action names to sentences, \ie, action definitions, which reduced the ambiguity of short action names. However these action definitions still lack rich contextual descriptions regarding the evolution of visual concepts in videos. Our work proposes to use video descriptions (\ie, the ActionHub dataset) to help the model learn the semantics of actions. In contrast to previously used additional text data, our automatically collected ActionHub dataset is low-cost and easy-scalable, and provides rich textual information to describe actions in videos.

\begin{table*}[!ht]
\begin{center}
\caption{The comparison between the proposed ActionHub dataset and existing video datasets. The first section presents the video caption datasets collected from the open domain. The second section presents various action video datasets, in which the first part shows various action datasets without captions, and the second part of this section presents the existing action video caption datasets.}
\label{tab:video_dataset_comparison}
\resizebox{\textwidth}{!}{% so there is no overfull
\begin{tabular}{c@{\hspace{0em}} |!{\vline} c c c c c c c c c}
\toprule
\multicolumn{2}{c}{\textbf{~~~~~~~~~~~~~~~Dataset}} & \textbf{Domain} & \textbf{\makecell[c]{Videos \\ (Clips)}} & \textbf{\makecell[c]{Video \\ Sources}} & \textbf{Captions} & \textbf{Sentences} & \textbf{\makecell[c]{Actions \\ (Steps)}} & \textbf{\makecell[c]{Caption/Class \\ Collection}} & \textbf{Year} \\
\toprule 
\multirow{16}{*}{\rotatebox{90}{\textbf{\makecell[c]{Video Caption Datasets \\ in Open-domain}}}}
~ & MSVD~\cite{chen-dolan-2011-collecting} & open & 2K  & YouTube & 70K & 70K & N/A & manually & 2011 \\
~ & M-VAD~\cite{torabi2015using} & open & 92 (49K) & Movies & 56K & 56K & N/A & automatically & 2015 \\
~ & MPII-MD~\cite{7298940} & open & 94 (69K) & Movies & 69K & 69K & N/A & automatically & 2015 \\
~ & TGIF~\cite{li2016tgif} & open & 102K & Tumblr & 126K & 126K & N/A & manually & 2016 \\
~ & MSR-VTT~\cite{xu2016msr} & open & 10K & YouTube & 200K & 200K & N/A & manually & 2016 \\
~ & DiDeMo~\cite{anne2017localizing} & open & 10.4K (26.8K) & Flickr & 40.5K & 40.5K & N/A & automatically & 2017 \\
~ & LSMDC~\cite{Rohrbach2016MovieD} & open & 200 (118K) & Movies & 128K & 128K & N/A & automatically & 2017 \\
~ & How2~\cite{sanabria2018how2} & open & 13K (185K) & YouTube & 185K & 185K & N/A & automatically & 2018 \\
~ & VATEX~\cite{wang2019vatex} & open & 41.3K & Kinetics-600 & 826K & 826K & 600 & manually & 2019 \\
~ & TRECVID-2020~\cite{awad2021trecvid} & open & 9.1K & Flickr & 28K & N/A & N/A & manually & 2020 \\
~ & TVR~\cite{lei2020tvr} & open & 21.8K & TV shows & 109K & N/A & N/A & manually & 2020 \\
~ & VIOLIN~\cite{liu2020violin} & open & 15.8K & TV shows & 95.3K & N/A & N/A & manually & 2020 \\
~ & WebVid-2M~\cite{bain2021frozen} & open & 2.5M & Footage Websites & 2.5M & 2.5M & N/A & automatically & 2021 \\
~ & VideoCC3M~\cite{nagrani2022learning} & open & 6.3M & Internet & 970K & N/A & N/A & automatically & 2022 \\
~ & \textbf{\blue{ActionHub}} & \textcolor{red}{\textbf{action}} & \textbf{\blue{3.6M}} & \textbf{\blue{Footage Websites}} & \textbf{\blue{3.6M}} & \textbf{\blue{10.1M}} & \textbf{\blue{1211}} & \textbf{\blue{automatically}} & \textbf{\blue{Now}} \\
%~ & xxx & domain & num.video & video.source & num.caption & num.sentence & num.class & manually/automatically & year \\
\toprule
\addlinespace[0.5ex]
\cmidrule{2-10}
\multirow{36}{*}{\rotatebox{90}{\makecell[c]{\textbf{Action Video Datasets}}}} & \multicolumn{6}{l}{\textbf{action datasets without captions}} & \\
\cdashline{2-10}
\addlinespace
~ & Olympic Sports~\cite{niebles2010modeling} & sport-action & 800 & YouTube & N/A & N/A & 16 & manually & 2010 \\
~ & HMDB51~\cite{hmdb51} & action & 7K & Movie etc. & N/A & N/A & 51 & manually & 2011 \\
~ & UCF101~\cite{ucf101} & action & 13.3K & YouTube & N/A &N/A  & 101 & manually & 2012 \\
~ & MPII Cooking~\cite{6247801} & cooking-action & 44 (5.6K) & manual-record & N/A & N/A & 65 & manually & 2012 \\
~ & Sports1M~\cite{6909619} & action & 1.1M & YouTube & N/A & N/A & 487 & automatically & 2014 \\
~ & Breakfast~\cite{Kuehne12} & cooking-action & 1.9K (8.4K) & manual-record & N/A & N/A & 10 (48) & manually & 2014 \\
~ & ActivityNet~\cite{caba2015activitynet} & action & 28K & YouTube & N/A  & N/A & 203 & manually & 2015 \\
~ & YouTube-8M~\cite{abu2016youtube} & action & 8M & YouTube & N/A & N/A & 4800 & automatically & 2016 \\
~ & Something V2~\cite{goyal2017something} & action & 220K & Manual-record & N/A & N/A & 174 & manually & 2017 \\
~ & FCVID~\cite{FCVID} & open & 91K & Internet & N/A & N/A & 239 & manually & 2017 \\
~ & AVA~\cite{gu2018ava} & atomic-action & 430 (385K) & YouTube & N/A & N/A & 80 & manually & 2017 \\
~ & MiT~\cite{monfort2019moments} & action & 1M & Internet & N/A & N/A & 339 & manually & 2018 \\
~ & Kinetics700~\cite{carreira2019short} & action & 650K & YouTube & N/A & N/A & 700 & manually & 2019 \\
~ & HACS Clips~\cite{zhao2019hacs} & action & 504K (1.5M) & YouTube & N/A & N/A & 200 & manually & 2019 \\
~ & HACS Segments~\cite{zhao2019hacs} & action & 50K (139M) & YouTube & N/A & N/A  & 200 & manually & 2019 \\
~ & COIN~\cite{tang2019coin} & instruction-action & 11.8K (46.3K) & YouTube & N/A & N/A & 180 (778) & manually & 2019 \\
% ~ & HVU & action & 572K & video.source & num.caption & num.sentence & 739 & manually & 2020 \\
~ & AViD~\cite{piergiovanni2020avid} & action & 450K & Flickr, etc. & N/A & N/A & 887 & manually & 2020 \\
~ & FineGym v1.0~\cite{shao2020finegym} & sport-action & 303 (32K) & Internet & N/A & N/A & 530 & manually & 2020 \\
~ & EPIC-Kitchens-100~\cite{damen2020rescaling} & cooking-action & 700 (76K) & Home &  N/A & N/A & \makecell[c]{97 verbs \\ 300 nouns} & manually & 2021 \\
%~ & EPIC-Kitchens-100 & cooking-action & 700 (76K) & Home & 19K & - & \makecell[c]{97 verbs \\ 300 nouns} & AMT workers & 2021 \\
~ & FineAction~\cite{liu2022fineaction} & daily-action & 16K (103K) & \makecell[c]{Existing datasets \\ \& Internet}  & N/A & N/A & 106 & manually & 2021 \\
~ & MultiSports~\cite{li2021multisports} & sport-action & 3.2K (37K) & YouTube & N/A & N/A & 66 & manually & 2021 \\
~ & UnlabeledHybrid~\cite{wang2023videomae} & action & 1.3M & \makecell[c]{Existing datasets \\ \& Instagram} & N/A & N/A & N/A & automatically & 2023 \\
~ & labeledHybrid~\cite{wang2023videomae} & action & 0.66M & Kinetics & N/A & N/A & 710 & automatically & 2023 \\
~ & \textbf{\blue{ActionHub}} & \textbf{\blue{action}} & \textbf{\blue{3.6M}} & \textbf{\blue{Footage Websites}} & \textcolor{red}{\textbf{3.6M}} & \textcolor{red}{\textbf{10.1M}} & \textbf{\blue{1211}} & \textbf{\blue{automatically}} & \textbf{\blue{Now}} \\
%~ & xxx & domain & num.video & video.source & num.caption & num.sentence & num.class & manually/automatically & year \\
\cmidrule{2-10}
\addlinespace[0.5ex]
\cmidrule{2-10}
~ & \multicolumn{6}{l}{\textbf{action datasets with captions}} & \\
\cdashline{2-10}
\addlinespace
~ & TACoS~\cite{regneri-etal-2013-grounding} & cooking-action & 127 (3.5K) & MPII Cooking & 11.8K & 11.8K & 26 & manually & 2013 \\
~ & TACoS-MLevel~\cite{rohrbach2014coherent} & cooking-action & 185 (25K) & MPII Cooking & 75K & 75K & 67 & manually & 2014 \\
~ & Charades~\cite{sigurdsson2016hollywood} & indoor-action & 9.8K (66K) &  Manual-record & 27K & - & 157 & manually & 2016\\
~ & ActivityNet-Cap~\cite{krishna2017dense} & daily-action & 20K & Internet & 100K & 100K & 200 & manually & 2017 \\
~ & YouCook2~\cite{zhou2018towards} & cooking-action & 2K (13.8K) & YouTube & 13.8K & 13.8K & 89 & manually & 2018 \\
~ & HowTo100M~\cite{miech2019howto100m} & instruction-action & 136M & YouTube & 136M & 136M &N /A  & automatically & 2019 \\
~ & How2R~\cite{li2020hero} & instruction-action & 9.3K & HowTo100M & 51K & N/A & N/A & manually & 2020 \\
~ & Ego4D~\cite{grauman2022ego4d} & ego-action & 3670 hours & manual-record & N/A & 3.85M & \makecell[c]{1772 verbs \\ 4336 nouns} & manually & 2021 \\
~ & Kinetic-GEB+~\cite{wang2022geb+} & action & 12K & Kinetics-400 & 177K & N/A & 400 & manually & 2022 \\
~ & \textbf{\blue{ActionHub}} & \textbf{\blue{action}} & \textbf{\blue{3.6M}} & \textbf{\blue{Footage Websites}} & \textbf{\blue{3.6M}} & \textbf{\blue{10.1M}} & \textcolor{red}{\textbf{1211}} & \textbf{\blue{\makecell[c]{automatically}}} & \textbf{\blue{Now}} \\
%~ & xxx & domain & num.video & video.source & num.caption & num.sentence & num.class & manually/automatically & year \\
\cmidrule{2-10}
\bottomrule
\end{tabular}
}
\end{center}
\end{table*}

\section{The ActionHub Dataset}

In this section, we describe our proposed action video description dataset, called \textbf{ActionHub}. We will first explain the motivation for collecting the ActionHub dataset, and then introduce the dataset. Later, we show some statistical analysis of the proposed ActionHub dataset.

\subsection{Motivation for ActionHub} 
\label{motivation_for_actionhub}
For zero-shot action recognition (ZSAR), learning a transferable alignment between video and text (\ie, class descriptions) data of seen actions is the premise of recognizing unseen actions. Existing ZSAR works~\cite{fu2014transductive, wang2017zero, hahn2019action2vec} mostly used class labels (\ie, action names) or attributes to extract semantic features. ER model~\cite{chen2021elaborative} additionally asked human annotators to provide precise action definitions based on sources like Wikipedia.
These approaches only consider simple phrases or one sentence to represent each action, which lack the textual context information to align the diverse video content, leading to a cross-modality diversity gap between video and text data in ZSAR.
In this work, we propose to utilize human-annotated video descriptions to enrich the diversity of class descriptions (\ie, text modality) of each action. 
However, existing action video caption datasets (\eg, Charades~\cite{sigurdsson2016hollywood} and Kinetic-GEB+~\cite{wang2022geb+}) fail to achieve this because of their limited number of actions. To this end, we propose the ActionHub dataset, which is the largest action video description dataset to date.

Table~\ref{tab:video_dataset_comparison} shows a full comparison between our ActionHub dataset and existing video datasets. The first section presents many large-scale video caption datasets, such as WebVid-2M~\cite{bain2021frozen} and VideoCC3M~\cite{nagrani2022learning}, while all of these video caption datasets are collected from the open domains (\ie, encompass a broad spectrum of visuals and topics), rather than the specific action domain. In contrast, our ActionHub dataset is automatically collected from action videos on public footage websites, which provides rich content descriptions to help models understand human actions. The second section in Table~\ref{tab:video_dataset_comparison} presents various video datasets in the action domain. 
The first part of this section shows many popular action datasets, in which some datasets cover a large number of common actions (\eg, 700 actions in Kinetics-700~\cite{carreira2019short} and 887 actions in AViD~\cite{piergiovanni2020avid}). However, all these action datasets lack video descriptions to describe the actions in videos. The second part presents the existing action video caption datasets (\eg, Ego4D~\cite{grauman2022ego4d} and Kinetic-GEB+~\cite{wang2022geb+}). Compared to all of these action video caption datasets, our ActionHub dataset shows superiority in the following aspects. 
\textit{i. Number of actions:} our ActionHub covers 1211 common actions, which is well-defined and largely exceeds the number of actions in existing action caption datasets; \textit{ii. Diversity of actions:} most existing action caption datasets are limited to specific domains (\eg, Charades~\cite{sigurdsson2016hollywood} is dominated by indoor-actions, YouCook2~\cite{zhou2018towards} and HowTo100M~\cite{miech2019howto100m} are dominated by cooking-actions), while our ActionHub avoids this problem by building actions based on several existing action datasets in various action domains; \textit{iii. Scalability:} the captions in existing action caption datasets are generally annotated by the hired annotators. However, our ActionHub automatically collects the descriptions provided by the users of public footage websites, which makes the collection of ActionHub easy-scalable and low-cost; \textit{iv. Semantics of descriptions:} captions provided by some action caption datasets have noisy semantics (\eg, the captions in HowTo100M~\cite{miech2019howto100m} are weakly or even incorrectly paired with videos). While the descriptions in ActionHub collected from footage websites can provide rich contextual descriptions of actions (see examples in Figure~\ref{examples_for_ActionHub}).

\subsection{Data Collection} 
To collect a diverse action video description dataset for zero-shot action recognition, we build large-scale video queries using action names from existing seven action recognition datasets, which include Kinetics-700~\cite{carreira2019short}, Moments-in-Time~\cite{monfort2019moments}, AVA~\cite{gu2018ava}, ActivityNet~\cite{caba2015activitynet} Olympic Sports~\cite{niebles2010modeling}, HMDB51~\cite{hmdb51}, and UCF101~\cite{ucf101}. 
As shown in Figure~\ref{collection_of_ActionHub}, given a total of 1490 actions from seven popular video action datasets,
we apply stop-word removal~\cite{DBLP:books/cu/LeskovecRU14} and word lemmatization~\cite{plisson2004rule} to the action names.
We then remove duplicates and end up with 1211 actions.
Each of the original action names is used as a search query to retrieve top-relevant videos from a video footage website, constraining that the videos must contain at least one person.
% top-1000 relevant videos?
For each video in the returned video list of an action query, we store the video description provided by the website user when uploading the video to the website.
As shown in Figure~\ref{collection_of_ActionHub}, the video descriptions contain much more diverse content information compared to Wikipedia-style action definitions used in ER model~\cite{chen2021elaborative}.
%Note that these captions are generated by human annotators specific to each video.

\begin{figure}[!t]
\centering
\includegraphics[width=1.0\columnwidth]{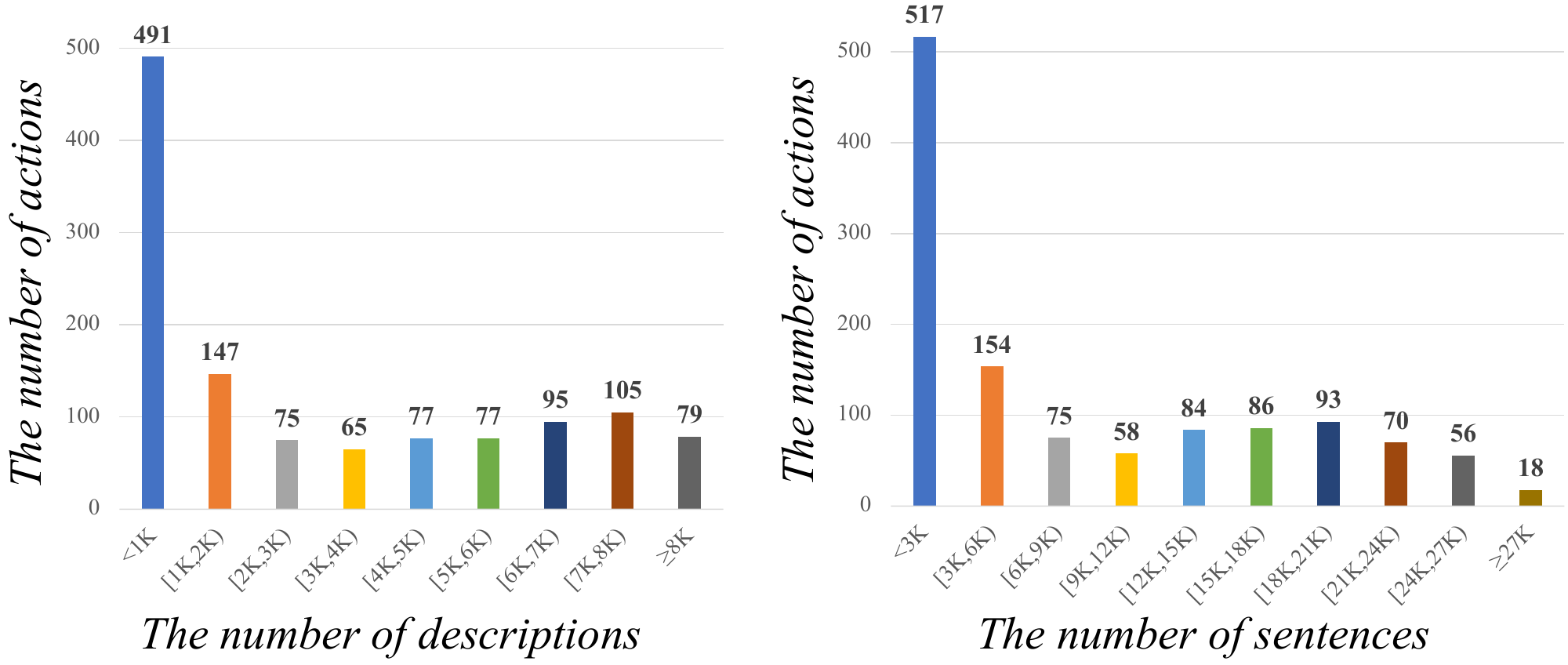}
\caption{Left: the statistics on the number of actions with respect to different numbers of video descriptions. Right: the statistics on the number of actions with respect to different numbers of sentences. In the collected ActionHub dataset, nearly half of the action classes have less than 1,000 video descriptions and 3,000 sentences. (Zoom in for a better view)}
\label{stats_actions_with_caps_sents}
\end{figure}

\begin{figure}[!t]
\centering
\includegraphics[width=1.0\columnwidth]{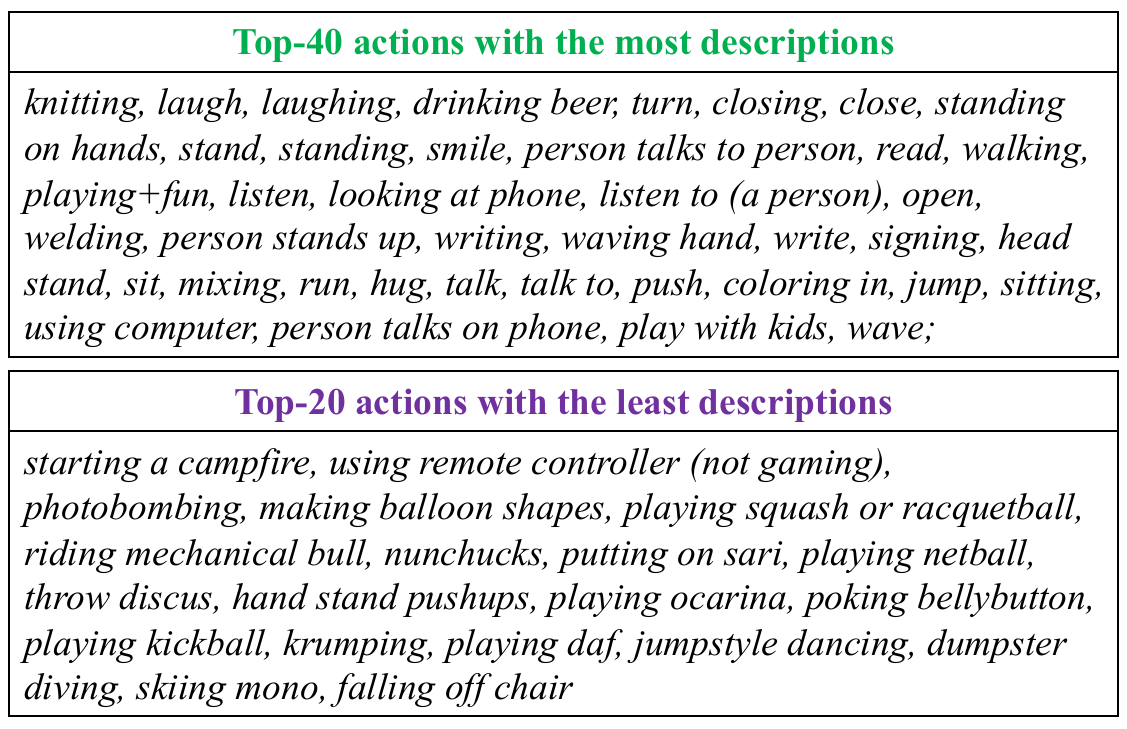}
\caption{The top-40 actions with the most descriptions and top-20 actions with the least descriptions, respectively. The actions with the most descriptions are more common in our daily life. And the actions with the least descriptions are rarely seen in real scenarios.}
\label{stats_actions_most_least}
\end{figure}

\begin{figure}[!t]
  \centering
  \includegraphics[width=0.9\columnwidth]{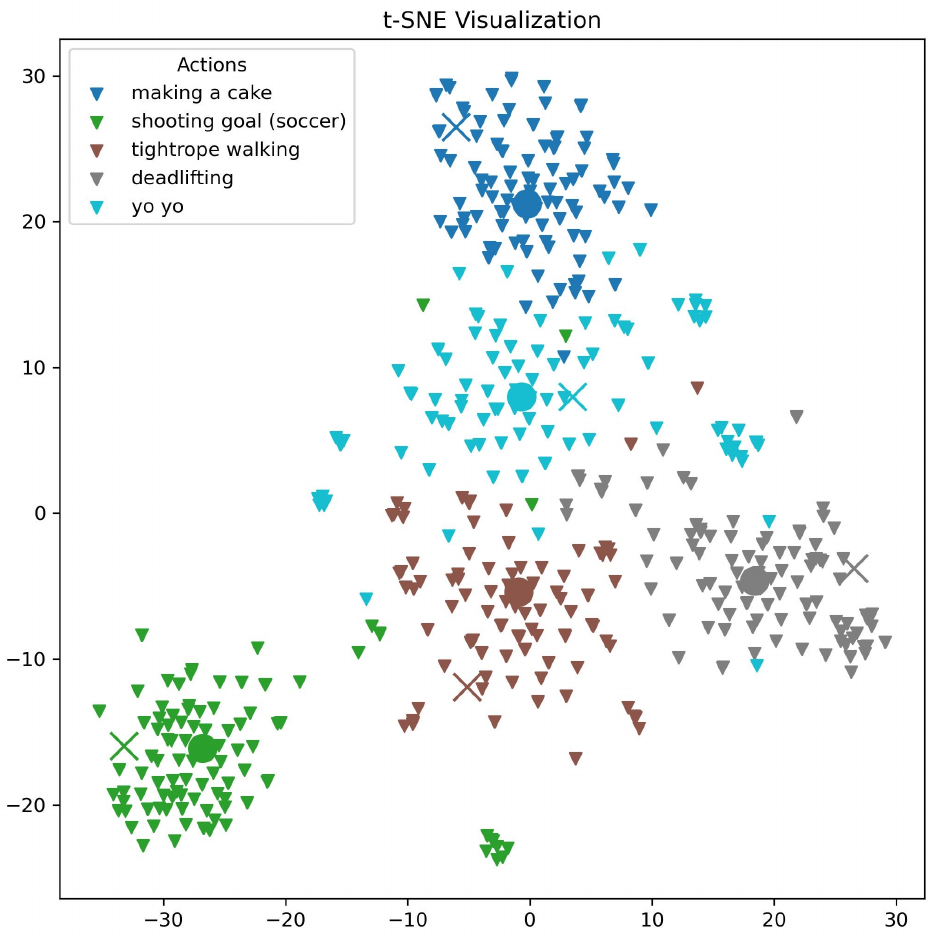} \\
  \includegraphics[width=1.0\columnwidth]{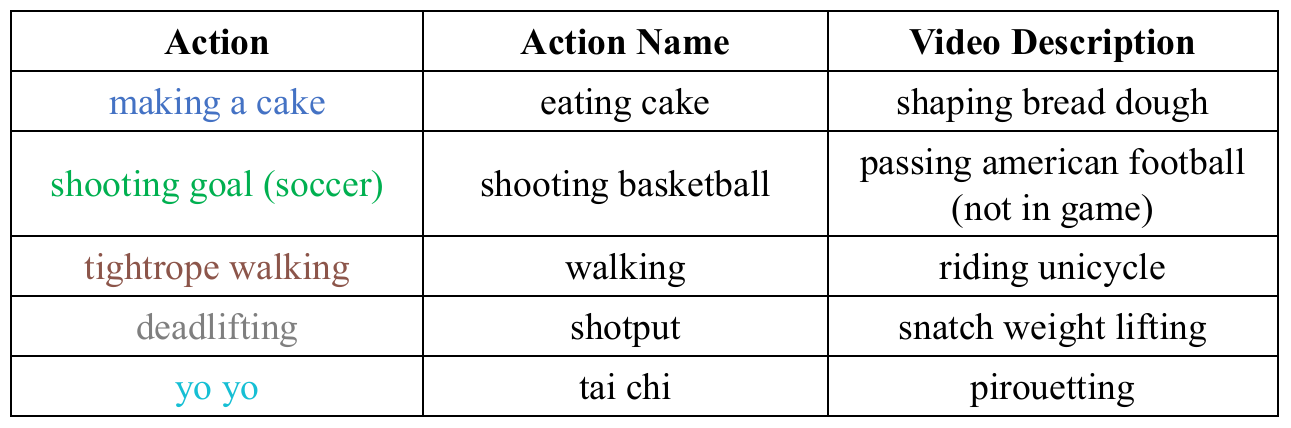}
  \caption{Visualizations of five action classes and their semantic correlations. The top part illustrates five action classes with different colors, using the averaged word2vec features of action names ($\times$), video descriptions ($\blacktriangledown$), and their centers ($\bullet$). The bottom part shows these five actions and identifies the most semantically correlated actions based on feature cosine similarity, for both action names and video descriptions. The video descriptions provide richer action-related semantics compared to action names alone, enhancing the understanding of each action.}
\label{action_name_desc_vis_text}
\end{figure}

\subsection{Statistical Analysis}
% /some comments here/
% %/*** 这里能不能对数据作些更深刻的统计，比如类之间的交叉和相似度，有没有长尾分布问题，有没有一些特别容易和难的例子，它们比例多少？有没有一些模凌两可的？总之做更深入的分析***/

% 1) action-name tsn-e, action-definition tsn-e, video-description tsn-e
% 2) action-name: top-N most-similar pairs, video-description top-N most-similar pairs; then show with video-description, the similar action pairs are more reasonable

In total, we have collected 3.6 million action video descriptions for 1211 human actions, with a total of 10.1 million sentences (averaging 3 sentences per video description). See Table~\ref{tab:video_dataset_comparison} for a comparison between our dataset and other video datasets. Our ActionHub dataset is the \textbf{largest} action video dataset with human annotated descriptions to date. See Figure~\ref{examples_for_ActionHub} for samples of the dataset.

In Figure~\ref{stats_actions_with_caps_sents}, we present the statistics on the number of actions, with respect to different numbers of descriptions (Figure~\ref{stats_actions_with_caps_sents}. Left) and sentences (Figure~\ref{stats_actions_with_caps_sents}. Right). We find that in the collected ActionHub dataset, nearly half of the action classes have less than 1,000 descriptions and 3,000 sentences, resulting in imbalanced distributions of the dataset. All the remaining action classes have over 1,000 descriptions and 3,000 sentences, and the distributions between these actions and descriptions/sentences are balanced. Figure~\ref{stats_actions_most_least} shows the top-40 actions with the most descriptions and top-20 actions with the least descriptions in our ActionHub dataset. And we can see that the actions with the most descriptions (\eg, jump, play with kids) are more common in our daily life. While those actions with the least descriptions (\eg, starting a campfire, jumpstyle dancing) are rarely seen in real scenarios. The above also shows that the actions in the ActionHub dataset are diverse.

At the top of Figure~\ref{action_name_desc_vis_text} we present five action classes, where each color represents a different action. For each action, we visualize the averaged Word2Vec~\cite{mikolov2013efficient} features of the corresponding \textit{action name} (denoted by $\times$), $100$ instances of \textit{video descriptions} (denoted by $\blacktriangledown$), and the \textit{center of video descriptions} (denoted by $\bullet$). The results demonstrate that the video descriptions in our ActionHub dataset exhibit action-related semantics with the corresponding actions in the feature space. And our video descriptions greatly enhance the textual diversity of each action, leading to improved alignment with video data. 
At the bottom of Figure~\ref{action_name_desc_vis_text}, we display the five actions. For each action, we determine the most semantically correlated actions based on the cosine similarity of their features, using both the features of the action name and the features of the center of video descriptions. We observe that the video descriptions provide more action-related semantics compared to the action names alone. For instance, for the action \textit{yo yo}, in terms of the feature similarity of action names, the most similar action is \textit{tai chi}. However, based on the feature similarity of video descriptions, the action $\textit{pirouetting}$ is a more reasonable choice as the most similar action.

\subsection{Impact of the ActionHub dataset}

The field of video understanding~\cite{wang2016temporal, wang2018non, lin2019tsm, feichtenhofer2019slowfast, bertasius2021space, Zhou_2021_CVPR, fan2021multiscale, liu2021video} has long been dedicated to analyzing human actions through the integration of video and language modalities. Recent advancements in vision-language models~\cite{ ma2023examination, gao2023clip, gan2022vision,
xu2021videoclip, wang2023all, lin2022frozen} involving video data in open domains (\ie, not specific to action videos), have demonstrated remarkable capabilities across various video understanding tasks. These models have greatly benefited from the availability of large-scale, paired video-text datasets, enabling them to acquire powerful abilities. However, the realm of videos in the action domain results in a notable gap in achieving this, \ie, there is a conspicuous absence of a large-scale multi-modal action video dataset. Existing large-scale video datasets fall short in several crucial dimensions: they do not cater specifically to videos in the action domain, they lack paired video-text data, they do not encompass a substantial volume of video samples, and they are deficient in the diversity of human actions. To address these limitations, this work introduces the ActionHub dataset, which covers a total of 1,211 distinct human actions and 3.6 million paired video-descriptions data. It is designed to fulfill the aforementioned criteria, providing the necessary foundation for the development of advanced video action understanding models. 

The proposed ActionHub dataset holds the potential to spark innovation and drive progress in multiple research domains. One particularly challenging area it addresses is the development of video-language foundation models in human action domain. This challenge lies in the previously mentioned data limitations. However, ActionHub offers a substantial collection of paired human action video-text data, poised to significantly advance the field. Furthermore, ActionHub stands out for its extensive coverage of human action classes, encompassing a diverse array of data domains. This broad spectrum of data can serve as a valuable resource for future research endeavors. Researchers can explore various dimensions of model generalization, including open-vocabulary learning and domain generalization, within the context of human action understanding. In this work, our investigation focuses on the zero-shot action recognition task, where we comprehensively show the effectiveness of the ActionHub dataset.

\vspace{0.3cm}

\section{Cross-modality and Cross-action Framework based on ActionHub}

In this work, we propose a \textbf{C}ross-m\textbf{o}dality and \textbf{C}ross-acti\textbf{o}n Modeling (\textbf{CoCo}) framework for ZSAR. By utilizing the rich textual descriptions of actions in our ActionHub dataset, the cross-modality diversity gap between video and text data of actions is alleviated, thus providing a prerequisite for the proposed CoCo framework to learn an effective alignment for recognizing unseen actions. In the following, we first describe our problem setting and give an overview of the proposed CoCo framework. Next, each module of our CoCo framework is elaborated.

\begin{figure*}[!t]
\centering
\includegraphics[width=1.0\textwidth]{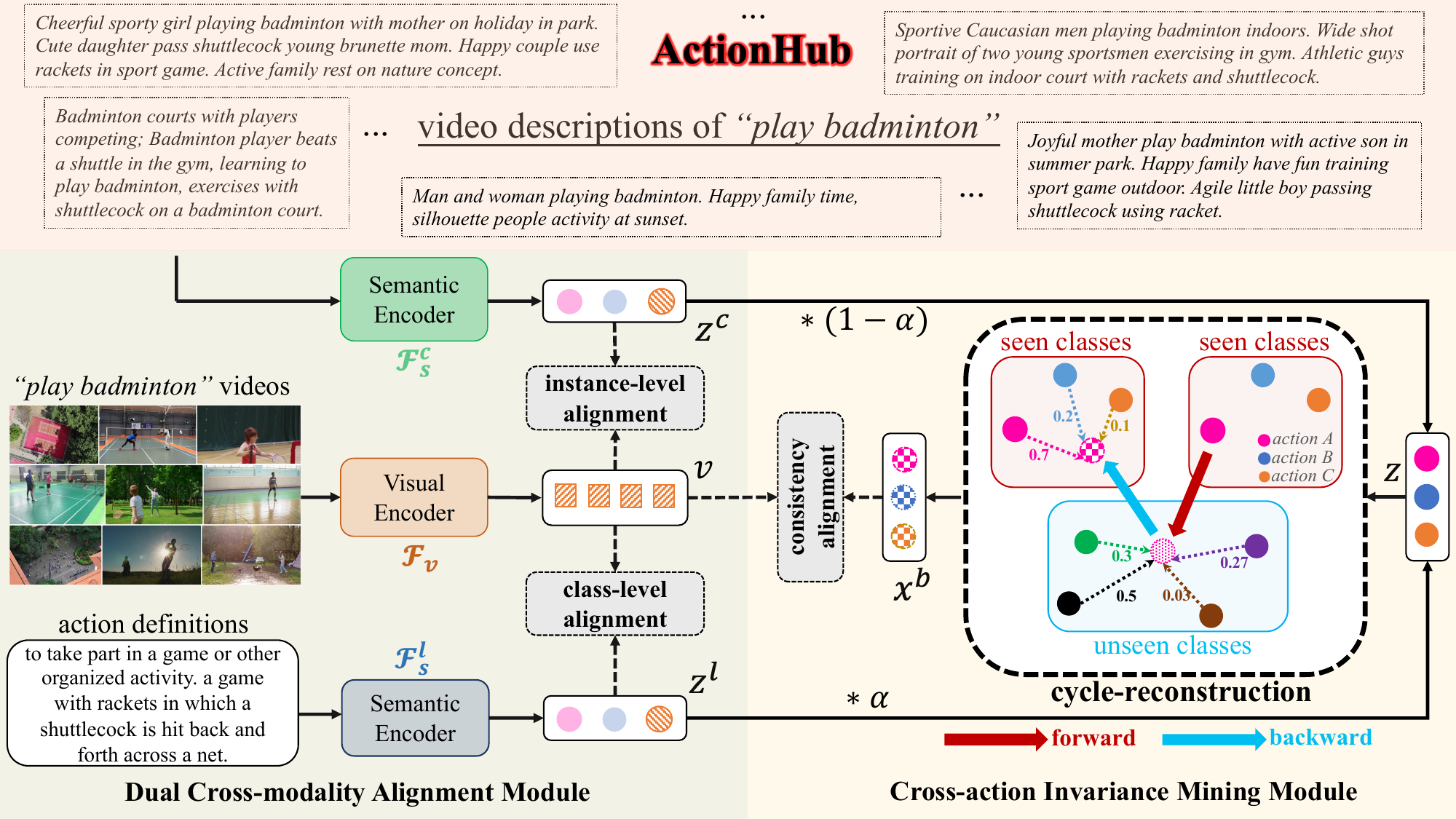}
\caption{An overview of the proposed CoCo framework, which consists of a Dual Cross-modality Alignment module and a Cross-action Invariance Mining module. The Dual Cross-modality Alignment module utilizes both action definition and video descriptions from the ActionHub dataset to enrich the diversity of class descriptions of each action. Then alignments between the video features and the class semantic features of each action are learned. The Cross-action Invariance Mining module exploits a cycle-reconstruction process on the class semantic feature spaces between seen actions and unseen actions, aiming to model the cross-action invariance by constraining the semantic consistency of the reconstructed class semantic features.}
\label{model}
\end{figure*}

\subsection{Problem Formulation}
Assume we have a training set including $N^s$ labeled videos $\mathcal{D}^s=\{V^s_i,Y^s_i\}_{i=1}^{N^s}$ with $\mathcal{C}^s$ seen actions, where $V^s_i\in V^s$ is the $i$-th video instance, and $Y^s_i\in Y^s$ is the corresponding label of the video. In addition, we have a test set including $N^u$ labeled videos $\mathcal{D}^u=\{V^u_i,Y^u_i\}_{i=1}^{N^u}$ with $\mathcal{C}^u$ unseen actions. Here, $V^s\cup V^u=V$, $Y^s\cup Y^u=Y$, and the actions in the training set and test set are disjoint, \ie, $Y^s\cap Y^u=\emptyset$. Zero-shot action recognition (ZSAR) aims to train an alignment model on the data $\{V^s,Y\}$, thus it can generalize to the unseen videos $V^u$ for zero-shot action classification. The superscripts $s/u$ that distinguish the seen/unseen data are omitted for a clearer representation in the following. 

In addition, we denote the ActionHub dataset as $\mathcal{A}=\{\{S_j^m, D_j^m\}_{j=1}^{N_m}\}_{m=1}^M$, where $M$ is the number of actions, $N_m$ is the number of video descriptions of the $m$-th action. $S_j^m$ and $D_j^m$ are the $j$-th video and video description of the $m$-th action, respectively. 

\subsection{Overview}
As shown in Figure~\ref{model}, our CoCo framework consists of a Dual Cross-modality Alignment module and a Cross-action Invariance Mining module.
The Dual Cross-modality Alignment module utilizes both action definitions and various video descriptions from the proposed \textbf{ActionHub} dataset to enrich the diversity of action class descriptions, thus better class semantic features can be obtained for feature alignment.
The Cross-action Invariance Mining module exploits a cycle-reconstruction process between the class semantic feature spaces of seen actions and unseen actions, aiming to learn the cross-action invariant representations by constraining the semantic consistency of the cycle-reconstructed class semantic features.

\subsection{Dual Cross-modality Alignment Module}
Since the class descriptions used in existing works are usually short action names or simple action definitions, which lack the textual context information corresponding to videos with diverse content, the learned visual space and semantic space cannot be well aligned.
Therefore, we propose the Dual Cross-modality Alignment module, which consists of a class-level alignment and an instance-level alignment. The class-level alignment utilizes class-level action definitions to match the semantics of video data. In addition to learning such class-level semantic correspondence, the instance-level alignment further utilizes various video description instances from the \textbf{ActionHub} dataset, aiming to match the semantics of visual context in videos. In this way, the diversity of class descriptions for each action is enriched, and a more generalizable cross-modality alignment is learned for recognizing unseen actions.

For the $i$-th video instance $V_i\in V$ of class $Y_i$, we first evenly sample $T$ segments $\hat{V}_i\in \mathbb{R}^{T\times 3 \times H\times W}$ from the video, and then use a visual encoder $\mathcal{F}_v$ to model its spatio-temporal feature $v_i$:
\begin{align}
v_i=\mathcal{F}_v(\hat{V}_i)\in \mathbb{R}^d,
\end{align}
where $d$ is the feature dimension. For the action definitions of $\mathcal{C}$ actions $Z^l=\{Z^l_1, Z^l_2, \cdots, Z^l_{\mathcal{C}}\}$, we use a semantic encoder $\mathcal{F}^l_s$ to model the action definition features $z^l$:
\begin{align}
z^l=\mathcal{F}^l_s(Z^l)=\{z^l_1, z^l_2, \cdots, z^l_{\mathcal{C}}\} \in \mathbb{R}^{\mathcal{C}\times d}.
\label{F_s_l}
\end{align}
Following the practice of existing works, we first learn a class-level alignment between the video feature $v_i$ and the corresponding class-level action definition feature $z^l_{Y_i}$, which is formulated as follows:
\begin{gather}
\mathcal{L}^l=-\sum_{i=1}^N\sum_{j=1}^{\mathcal{C}}\mathbb{I}(j=Y_i) \cdot log\frac{exp(p^l_{ij}/\tau)}{\sum_{k=1}^{\mathcal{C}}exp(p^l_{ik}/\tau)}, \\
p^l_{ij}= v_i \cdot z^l_j \in \mathbb{R}^1,
\end{gather}
where $\mathcal{L}^l$ is a standard contrastive loss. $p^l_{ij}$ represents the similarity between the $i$-th video feature $v_i$ and the action definition feature $z^l_j$ of action class $j$. $\mathbb{I}(\cdot)$ is an indicator function and $\tau$ is a temperature hyper-parameter.

However, the semantics of action definitions used above cannot sufficiently represent that of the videos, which causes misalignment between the learned visual space and semantic space.
Therefore, to make up for the semantic deficiency of the action definitions, the Dual Cross-modality Alignment module utilizes the video descriptions from the proposed ActionHub dataset to enrich the semantics of class descriptions for each action. 

To this end, for each action class $j$, we select top-$k$ instances of video descriptions in terms of their relevance to the action class, \ie, $D^{j}=\{D^{j}_{top-1}, D^{j}_{top-2}, \cdots, D^{j}_{top-k}\}$. Specifically, for each video description of class $j$, we obtain its feature embedding by averaging the Word2Vec features~\cite{mikolov2013efficient} of all words in the description. Then the relevance is calculated by cosine similarity between the feature embedding of the description and the Word2Vec feature of the action name of class $j$.
Later, we also use a semantic encoder $\mathcal{F}_s^c$ to extract their semantic features $d^{j}$:
\begin{align}
d^{j}=\mathcal{F}_s^c(D^{j})=\{d^{j}_1, d^{j}_2, \cdots, d^{j}_k\} \in \mathbb{R}^{k\times d}.
\label{F_s_c}
\end{align}
Next, we aggregate features of these video description instances to obtain action-representative content feature $z^c_j$ for each action class $j$:
\begin{align}
z^c_j=\frac{1}{k}\sum_{i=1}^{k}d^j_i \in \mathbb{R}^d.
\end{align}
Finally, we use the instance-based content feature $z^c_j$ to learn an instance-level alignment between the learned visual space and semantic space:
\begin{gather}
\mathcal{L}^c=-\sum_{i=1}^N\sum_{j=1}^{\mathcal{C}}\mathbb{I}(j=Y_i) \cdot log\frac{exp(p^c_{ij}/\tau)}{\sum_{k=1}^{\mathcal{C}}exp(p^c_{ik}/\tau)},\\
p^c_{ij}= v_i \cdot z^c_j \in \mathbb{R}^1,
\end{gather}
where $\mathcal{L}^c$ is also a standard contrastive loss. $p^c_{ij}$ represents the similarity between the $i$-th video feature $v_i$ and the content feature $z^c_j$ of action class $j$.

By utilizing both action definitions and video descriptions, our Dual Cross-modality Alignment module obtains rich class semantic features for each action, thus the alignment can be well established by:
\begin{gather}
\mathcal{L}^{align}=\mathcal{L}^l+\mathcal{L}^c,
\end{gather}
where $\mathcal{L}^{align}$ is an overall loss for learning the alignment between video feature space and class description feature space of actions.

\subsection{Cross-action Invariance Mining Module}
There are many common semantic concepts (\eg, human-object interactions, atomic actions) between actions, whose semantics are invariant and discriminative across seen actions and unseen actions. As the semantics of each action class are composed of these concepts, modeling the feature representations of these cross-action invariant concepts can further boost the alignment for each action class.
To this end, our proposed Cross-action Invariance Mining module utilizes the class semantic feature space of unseen actions to cycle-reconstruct the class semantic feature of each seen action, and constraints the cycle-reconstructed class semantic features of seen actions to maintain their original semantics with a cycle-consistency loss, thus the model can focus on modeling the cross-action invariant and discriminative feature representations.

The feature space of class descriptions with rich semantics is the prerequisite for learning invariant semantics between actions. Thus we first fuse the action definition feature $z^l_j$ and content feature $z^c_j$ to obtain the class semantic feature $z_j$ with rich semantics for each action class $j$:
\begin{align}
\label{ED.SS.fusion1}
z_j=\alpha * z^l_j + (1-\alpha) * z^c_j,
\end{align}
where $\alpha\in (0,1)$ is a hyper-parameter that balances the semantics of the action definition feature and content feature. 

To cycle-reconstruct the class semantic feature of each seen action using the class semantic feature space of unseen actions, we exploit a cycle-reconstruction process, which consists of a forward pass and a backward pass. 
For the forward pass, we reconstruct the class semantic feature of each seen action $z^s_j\in \mathbb{R}^d$ from the class semantic feature space of $M$ unseen actions $z^u=\{z^u_1, z^u_2, \cdots, z^u_{M}\}\in \mathbb{R}^{M\times d}$ in an attention-based manner, where the forward-feature $x^f_j \in \mathbb{R}^d$ for the seen action class $j$ is obtained by:
\begin{align}
x^f_j=\sum_{m=1}^{M} \frac{exp(z^s_j \cdot z^u_m /\tau)}{\sum_{k=1}^{M}exp(z^s_j \cdot z^u_k/\tau)} * z^u_m.
\end{align}
To avoid the impact of the semantic gap between seen actions and unseen actions, the backward-pass transfers the forward-feature $x^f_j$ of each seen action back to class semantic feature space of $N$ seen actions $z^s=\{z^s_1, z^s_2, \cdots, z^s_{N}\}\in \mathbb{R}^{N\times d}$, where the cycle-feature $x^b_j \in \mathbb{R}^d$ for the seen action class $j$ is obtained by:
\begin{align}
x^b_j=\sum_{n=1}^{N} \frac{exp(x^f_j \cdot z^s_n /\tau)}{\sum_{k=1}^{N}exp(x^f_j \cdot z^s_k/\tau)} * z^s_n.
\label{backward}
\end{align}
To ensure the invariant semantics between actions are modeled during the cycle-reconstruction process, our Cross-action Invariance Mining module constraints the cycle-feature $x^b_j$ of the seen action class $j$ to maintain its original semantics, \ie, the consistency alignment between the cycle-feature $x^b_j$ and the video feature $v_i$ of the action class $j$ is established:
\begin{gather}
\mathcal{L}^{cycle}=-\sum_{i=1}^N\sum_{k=1}^{\mathcal{C}}\mathbb{I}(k=j) \cdot log\frac{exp(p^{cycle}_{ik}/\tau)}{\sum_{c=1}^{\mathcal{C}}exp(p^{cycle}_{ic}/\tau)}, \\
p^{cycle}_{ik}= v_i \cdot x^b_k \in \mathbb{R}^1,
\end{gather}
where $\mathcal{L}^{cycle}$ is the cycle-consistency loss. $p^{cycle}_{ik}$ represents the similarity between the $i$-th video feature $v_i$ and the cycle-feature $x^b_k$ of class description of action class $k$. The alignment learned by the cycle-consistency loss $\mathcal{L}^{cycle}$ ensures that our CoCo framework models cross-action invariant semantics.

In training, we optimize our CoCo framework by minimizing the following loss:
\begin{align}
\mathcal{L}^{overall}=\mathcal{L}^{align}+\gamma * \mathcal{L}^{cycle},
\end{align}
where $\gamma$ is a balance factor. In inference, for an unseen video $V_i$, our model predicts its label to be:
\begin{align}
\hat{Y_i}=\argmax_{j\in Y^u} \ v_i \cdot p_j, \\
\label{ED.SS.fusion2}
p_j = \alpha * z^l_j + (1-\alpha) * z^c_j,
\end{align}
where $v_i$ is the video feature, $p_j$ is the class semantic feature of unseen action class $j$ and $\alpha$ is the hyper-parameter.

\section{Experiments}

In this section, we first introduce the different learning protocols and popular benchmark datasets for ZSAR. Then the implementation details of the CoCo framework are elaborated. Later, we compare our method with existing works. Finally, both ablation studies and qualitative results are presented to demonstrate the effectiveness of the proposed ActionHub dataset and CoCo framework.

\subsection{Learning Protocols and Datasets for ZSAR}
There are two different learning protocols for ZSAR, namely, the intra-dataset protocol~\cite{xu2017transductive} and the cross-dataset protocol~\cite{brattoli2020rethinking}. Each learning protocol has different experimental settings on existing ZSAR benchmark datasets (\ie, Kinetics~\cite{carreira2017quo}, UCF101~\cite{ucf101}, and HMDB51~\cite{hmdb51}). 
All existing ZSAR works evaluate the performance of their models under only one protocol, in contrast, our work makes a comprehensive comparison using the two protocols for the first time.

\subsubsection{Intra-dataset Protocol}

Most existing ZSAR works~\cite{liu2011recognizing, xu2017transductive, jain2015objects2action, gao2019know, chen2021elaborative} follow the intra-dataset learning protocol, which samples seen actions (for training) and unseen actions (for test) from the same action video dataset. Under this protocol, three benchmark datasets, \ie, Kinetics-ZSAR~\cite{chen2021elaborative}, UCF101~\cite{ucf101}, and HMDB51~\cite{hmdb51} are used, and the training and test are conducted within each dataset.

\vspace{0.1cm}

\noindent \textbf{Kinetics-ZSAR.} The Kinetics-ZSAR benchmark is introduced by~\cite{chen2021elaborative}, which is currently the largest ZSAR dataset. It uses 400 action classes in Kinetics-400~\cite{carreira2017quo} as seen classes for training, and 220 new action classes in Kinetics-600~\cite{carreira2018short} that are not in Kinetics-400 as unseen classes for validation and test. 
The 220 new action classes are randomly split into 60 validation classes and 160 test classes, and these classes are independently split three times for repeated experiments. 
The Kinetics-ZSAR benchmark contains 212577 training videos, 2682 validation videos, and 14125 testing videos on average of the three splits.
We report the average top-$1$ and top-$5$ accuracies and standard deviations of the three splits on this dataset.

\vspace{0.1cm}

\noindent \textbf{UCF101 and HMDB51.} UCF101 contains 13320 video clips distributed among 101 action classes. 
%Each class has at least 100 video clips and each clip lasts an average duration of 7.2s. 
HMDB51 includes 6766 videos of 51 action classes. 
%Each action class has at least 101 video clips and each clip lasts an average duration of 4.3s. 
Under the intra-dataset protocol, for each dataset, we adopt 50 independent splits (50\% of actions are randomly selected for training and the rest are considered unseen actions for test in each split) provided by ~\cite{xu2017transductive}, and report the average accuracy and standard deviation.

\subsubsection{Cross-dataset Protocol}

Recently, several ZSAR works~\cite{brattoli2020rethinking, pu2022alignment, lin2022cross} applied a cross-dataset learning protocol in ZSAR, which samples seen actions from a large-scale action dataset (\eg, Kinetics-400~\cite{kay2017kinetics}) for training, and samples unseen actions from other action datasets (\eg, UCF101) for testing. Following~\cite{brattoli2020rethinking}, we use Kinetics-664 as the training set (obtained from Kinetics700~\cite{carreira2019short} with class filtering to avoid classes overlapping in UCF101 and HMDB51), and half of the classes from UCF101 and HMDB51 (50 classes for UCF101 and 25 classes for HMDB51) as the test sets. We repeat the evaluation ten times and report the average accuracy on each test dataset.

\vspace{0.1cm}

\OK{\subsection{Implementation Details}}
Our CoCo framework is implemented based on the ER model~\cite{chen2021elaborative}, which uses action definitions as class descriptions and elaborative rehearsal loss based on pre-detected objects to improve the generalization of the model. In CoCo, both the visual encoder $\mathcal{F}_v$ and semantic encoder $\mathcal{F}^l_s/\mathcal{F}^c_s$ are composed of a frozen backbone network for feature extraction and a learnable fully-connected layer for feature tuning. The semantic backbone in the semantic encoder $\mathcal{F}^l_s/\mathcal{F}^c_s$ is a pre-trained $12$-layer BERT~\cite{devlin2018bert}. To make our model more efficient, we share the parameters between the semantic encoders $\mathcal{F}^l_s$ and $\mathcal{F}^c_s$ (See Eq.~\ref{F_s_l} and Eq.~\ref{F_s_c}).

For the intra-dataset protocol, we follow the same experimental setting as ER~\cite{chen2021elaborative}. For Kinetics-ZSAR, the number of segments $T$ is set to $8$ and the visual backbone is TSM~\cite{lin2019tsm}. And for UCF101 and HMDB51, the number of segments $T$ is $1$ and the visual backbone is BiT~\cite{kolesnikov2020big}. For the cross-dataset protocol, we follow the same experimental setting as E2E-ZSAR~\cite{brattoli2020rethinking}, which uses R(2+1)D~\cite{tran2018closer} network as the visual backbone, and samples a single clip with $16$ frames and a stride of $4$ from each video for training and test. For both intra-dataset protocol and cross-dataset protocol, we do not fine-tune the backbones in visual and semantic encoders for fast training.

The feature dimension $d$ is $512$. The number of video descriptions $k$ selected from ActionHub for each action is $100$. The hyper-parameters $\tau$, $\alpha$, and $\gamma$ are set to $0.1$, $0.5$, and $0.1$, respectively. We use Adam to train the model with a weight decay of $1e-4$. The batch size is set to $512$. We set the learning rate as $1e-3$ and use warm-up and cosine annealing policies to adjust the learning rate. Our CoCo framework is implemented using PyTorch. All experiments are conducted on one Nvidia Tesla V100 GPU.

\vspace{0.3cm}

\subsection{Comparisons with the State-of-the-arts}
For intra-dataset protocol, we compare our CoCo with existing methods on Kinetics-ZSAR, HMDB51 and UCF101 datasets. Since all existing works~\cite{frome2013devise, akata2015label, akata2015evaluation, zhang2017learning, romera2015embarrassingly, ghosh2020all} using intra-dataset protocol do not train their backbone networks, for a fair comparison, we re-evaluate the ER model~\cite{chen2021elaborative} without training the BERT backbone using its official codebase. The result we obtained on Kinetics-ZSAR is consistent with that reported in the supplementary material of~\cite{chen2021elaborative}. As shown in Table~\ref{tab:comparisons_on_kinetics}, on the existing largest ZSAR benchmark dataset, \ie, Kinetics-ZSAR, our CoCo achieves $42.0\%$ in top-$1$ accuracy and $71.0\%$ in top-$5$ accuracy, which outperforms the state-of-the-art by $6.2\%$ in top-$1$ accuracy and $6.8\%$ in top-$5$ accuracy, respectively. For HMDB51 and UCF101, as shown in the \textit{Intra-dataset protocol} section of Table~\ref{tab:comparisons_on_ucf_hmdb}, our CoCo achieves $33.4\%$ and $51.2\%$ in top-$1$ accuracy on HMDB51 and UCF101 respectively, achieving state-of-the-art performance.

For the cross-dataset protocol, following existing practice, the model is trained on the Kinetics-664 dataset and tested on the HMDB51 and UCF101 datasets. All existing works~\cite{brattoli2020rethinking, lin2022cross, pu2022alignment} that use the cross-dataset protocol train their backbone networks end-to-end for more powerful feature representation, which consumes a lot of resources for training. To reduce the resource consumption, our model does not train the backbone networks. As shown in the \textit{Cross-dataset protocol} section of Table~\ref{tab:comparisons_on_ucf_hmdb}, our CoCo obtains $34.6\%$ and $57.6\%$ in top-$1$ accuracy on HMDB51 and UCF101 respectively, achieving comparable performance on HMDB51 and state-of-the-art on UCF101 without training the backbone networks, which further demonstrates the efficacy of our CoCo.

\begin{table}[!t]
%\tiny
\scriptsize
%\footnotesize
%\small
%\normalsize
%\large
\setlength\tabcolsep{9pt}
\caption{\textbf{Zero-shot action recognition results on Kinetics-ZSAR.} The results show that our CoCo framework significantly outperforms ER~\cite{chen2021elaborative} in both top-$1$ and top-$5$ accuracies.}
\begin{center}
\resizebox{\columnwidth}{!}{
\begin{tabular}{c c c}
\toprule
Methods & top-1 Acc (\%) & top-5 Acc (\%) \\
\hline
DEVISE~\cite{frome2013devise} & 23.8 $\pm$ 0.3 & 51.0 $\pm$ 0.6 \\
ALE~\cite{akata2015label} & 23.4 $\pm$ 0.8 & 50.3 $\pm$ 1.4 \\
SJE~\cite{akata2015evaluation} & 22.3 $\pm$ 0.6 & 48.2 $\pm$ 0.4 \\
DEM~\cite{zhang2017learning} & 23.6 $\pm$ 0.7 & 49.5 $\pm$ 0.4 \\
ESZSL~\cite{romera2015embarrassingly} & 22.9 $\pm$ 1.2 & 48.3 $\pm$ 0.8 \\
GCN~\cite{ghosh2020all} & 22.3 $\pm$ 0.6 & 49.7 $\pm$ 0.6 \\
ER~\cite{chen2021elaborative} & 35.8 $\pm$ 1.3 & 64.2 $\pm$ 1.2 \\ 
\hline
\textbf{Ours} &\textbf{42.0 $\pm$ 1.4} & \textbf{71.0 $\pm$ 0.5} \\
\bottomrule
\end{tabular}}
\end{center}
\label{tab:comparisons_on_kinetics}
\end{table}

\begin{table}
%\scriptsize
\normalsize
\setlength\tabcolsep{6pt}
\caption{\textbf{Zero-shot action recognition results on HMDB51 and UCF101.} For intra-dataset protocol, our model achieves state-of-the-art on both datasets. For cross-dataset protocol, our model achieves comparable performance on HMDB51 and state-of-the-art on UCF101 without training the backbone networks (the value of N in parentheses represents the number of clips used for testing).}
\begin{center}
%\resizebox{\columnwidth}{!}{
\begin{tabular}{c c c}
\toprule
Methods & HMDB51 & UCF101 \\
\toprule
\toprule 
\multicolumn{3}{c}{\textit{Intra-dataset protocol} (frozen backbone)} \\
\hdashline 
DAP~\cite{lampert2009learning} & N/A & 15.9 $\pm$ 1.2 \\
IAP~\cite{lampert2009learning} & N/A & 16.7 $\pm$ 1.1 \\
HAA~\cite{liu2011recognizing} & N/A & 14.9 $\pm$ 0.8 \\
SVE~\cite{xu2015semantic} & 13.0 $\pm$ 2.7 & 10.9 $\pm$ 1.5 \\
ESZSL~\cite{romera2015embarrassingly} & 18.5 $\pm$ 2.0 & 15.0 $\pm$ 1.3 \\
SJE-W~\cite{akata2015evaluation} & 13.3 $\pm$ 2.4 & 9.9 $\pm$ 1.4 \\
SJE-A~\cite{akata2015evaluation} & N/A & 12.0 $\pm$ 1.2 \\
MTE~\cite{xu2016multi} & 19.7 $\pm$ 1.6 & 15.8 $\pm$ 1.3 \\
ZSECOC~\cite{qin2017zero} & 22.6 $\pm$ 1.2 & 15.1 $\pm$ 1.7 \\
ETSAN~\cite{10093084} & 22.3 $\pm$ 2.4 & 20.6 $\pm$ 1.6 \\
UR~\cite{zhu2018towards} & 24.4 $\pm$ 1.6 & 17.5 $\pm$ 1.6 \\
O2A~\cite{jain2015objects2action} & 15.6 & 30.3 \\
ASR~\cite{wang2017alternative} & 21.8 $\pm$ 0.9 & 24.4 $\pm$ 1.0 \\
TS-GCN~\cite{gao2019know} & 23.2 $\pm$ 3.0 & 34.2 $\pm$ 3.1 \\
PS-GNN~\cite{9067001} & 32.6 $\pm$ N/A & 43.0 $\pm$ N/A \\
TSRL~\cite{zhuo2022zero} & 24.5 $\pm$ N/A & 48.9 $\pm$ N/A \\
ER~\cite{chen2021elaborative} & 29.3 $\pm$ 3.6 & 47.0 $\pm$ 2.8 \\
\textbf{Ours} & \textbf{33.4 $\pm$ 4.0} & \textbf{51.2 $\pm$ 2.9} \\
\bottomrule
\toprule
\multicolumn{3}{c}{\textit{Cross-dataset protocol} (tune backbone, except ours)} \\
\hdashline
E2E-ZSAR (N: 1)~\cite{brattoli2020rethinking} & 27.0 & 43.0 \\
E2E-ZSAR (N: 25)~\cite{brattoli2020rethinking} & 32.7 & 48.0 \\
PS-ZSAR (N: 25)~\cite{kerrigan2021reformulating} & 33.8 & 49.2 \\
ResT\_18 (N: 25)~\cite{lin2022cross} & N/A & 54.7 \\
AURL (N: 1)~\cite{pu2022alignment} & 34.3 & 55.1 \\
ViSET (N: 6)~\cite{Doshi_2023_CVPR} & 34.5 & 53.2 \\
\textbf{Ours (N: 1)} & \textbf{34.6} & \textbf{57.6} \\
\bottomrule
\end{tabular}
\end{center}
\label{tab:comparisons_on_ucf_hmdb}
\end{table}

\subsection{Ablation Studies}
In this section, we conduct experiments to demonstrate the effectiveness of each module of the proposed CoCo framework. 
Under the cross-dataset protocol, although our model achieves state-of-the-art performance without tuning backbones due to the limited resources, the capability of the model cannot be well explored. 
Therefore, all of the following experiments are conducted on the largest ZSAR dataset, \ie, Kinetics-ZSAR, under the more commonly used and low-cost intra-dataset protocol.

\vspace{0.1cm}

\noindent \textbf{- The effects of using different class descriptions.} 
We use the action definitions (\textit{AD-only}), video descriptions (\textit{VC-only}), and action definitions $+$ video descriptions (\textit{AD + VC}) to represent the class descriptions of actions, respectively. 
Table~\ref{tab:ablation_on_class_description} shows the effects of using different class descriptions on the Kinetics-ZSAR dataset. Here the Cross-action Invariance Mining module is not used to make the comparison of using different class descriptions clearer. 
The results in Table~\ref{tab:ablation_on_class_description} indicate that the performance of the \textit{VC-only} model is comparable to that of the \textit{AD-only} model, which means that the video descriptions from the ActionHub dataset have rich semantics and are discriminative for actions.
Moreover, the \textit{AD + VC} model can significantly improve the performance of ZSAR, which demonstrates that the video descriptions from ActionHub are complementary to existing action definitions and can effectively enrich the diversity of class descriptions (\ie, text modality) of actions. 

\begin{table}[h]
\normalsize
\setlength\tabcolsep{10pt}
\centering
\caption{The effects of using different class descriptions on the Kinetics-ZSAR dataset. Different class descriptions for actions, \ie, action definitions (AD-only), video descriptions (VC-only), and action definitions + video descriptions (AD + VC) are used, respectively.}
\begin{tabular}{c | c}
\hline
Class Descriptions & Acc (\%) \\
\hline
 AD-only & 35.8 \\
 VC-only & 36.1  \\
 AD + VC & \textbf{40.7}\\
\hline
\end{tabular}
\label{tab:ablation_on_class_description}
\end{table}

\vspace{0.1cm}

\begin{figure}[!h]
\centering
\includegraphics[width=0.8\columnwidth]{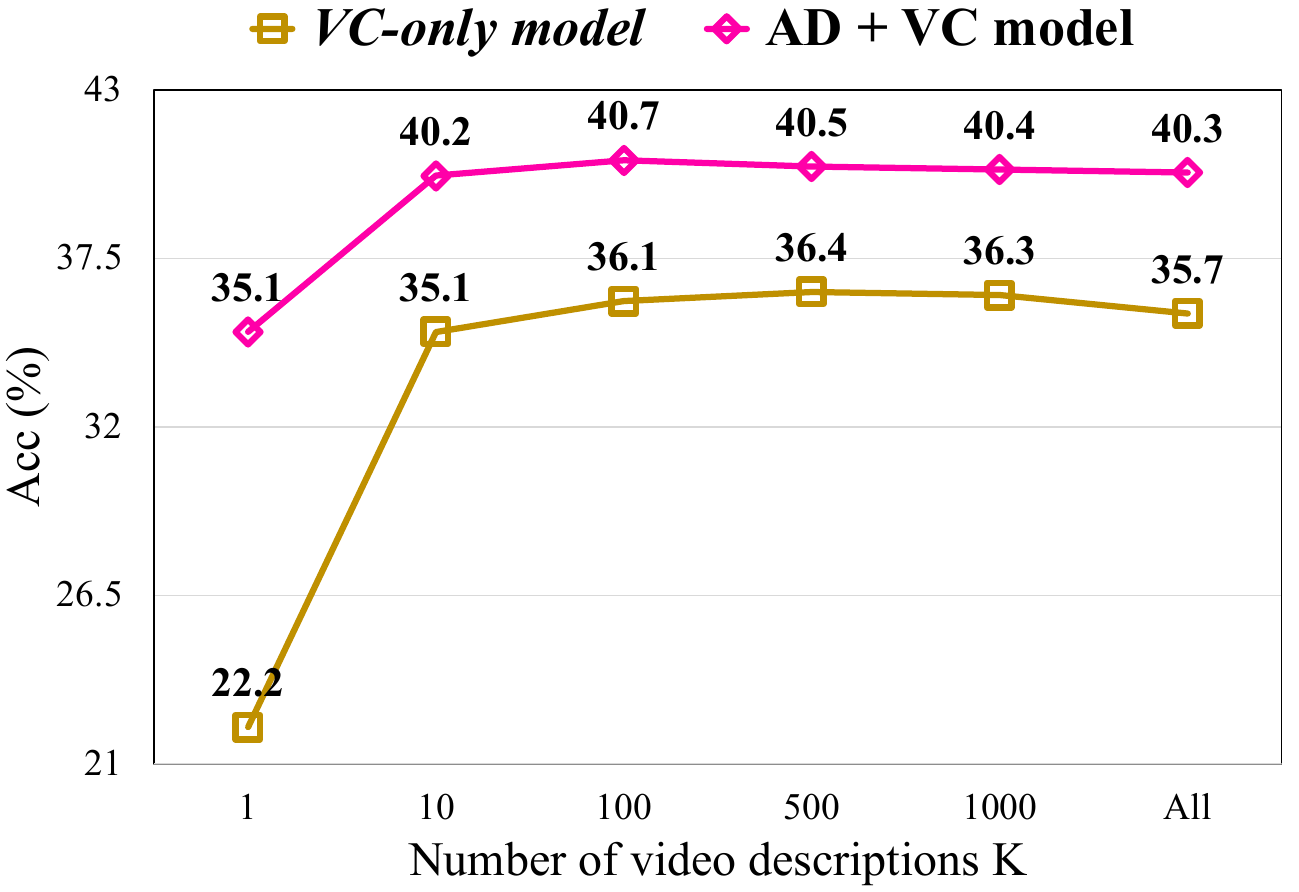}
\caption{The effects of the number of video descriptions $k$ selected from ActionHub. Our model achieves satisfactory performance on the Kinetics-ZSAR dataset when the top-100 relevant video descriptions are selected for each action.}
\label{fig:ablation_on_caption_numbers}
\end{figure}

\noindent \textbf{- The effects of the number of video descriptions $k$ selected from ActionHub.} For each action, we average features of its top-$k$ relevant video descriptions to obtain the content feature of action for feature alignment. As shown in Figure~\ref{fig:ablation_on_caption_numbers}, on the Kinetics-ZSAR dataset, the performance of both the \textit{VC-only} model and \textit{AD + VC} model can be continuously improved when we increase the number of video descriptions $k$ for each action. However, the performance of these two models will degrade sightly if $k$ becomes very large. The reason is that simply averaging the features of these video descriptions cannot effectively filter out the noise from video descriptions that are less relevant to the action. Based on this, it is a good trade-off to select the top-$100$ relevant video descriptions for each action.

\vspace{0.1cm}

\begin{figure}[!h]
\centering
\includegraphics[width=0.75\columnwidth]{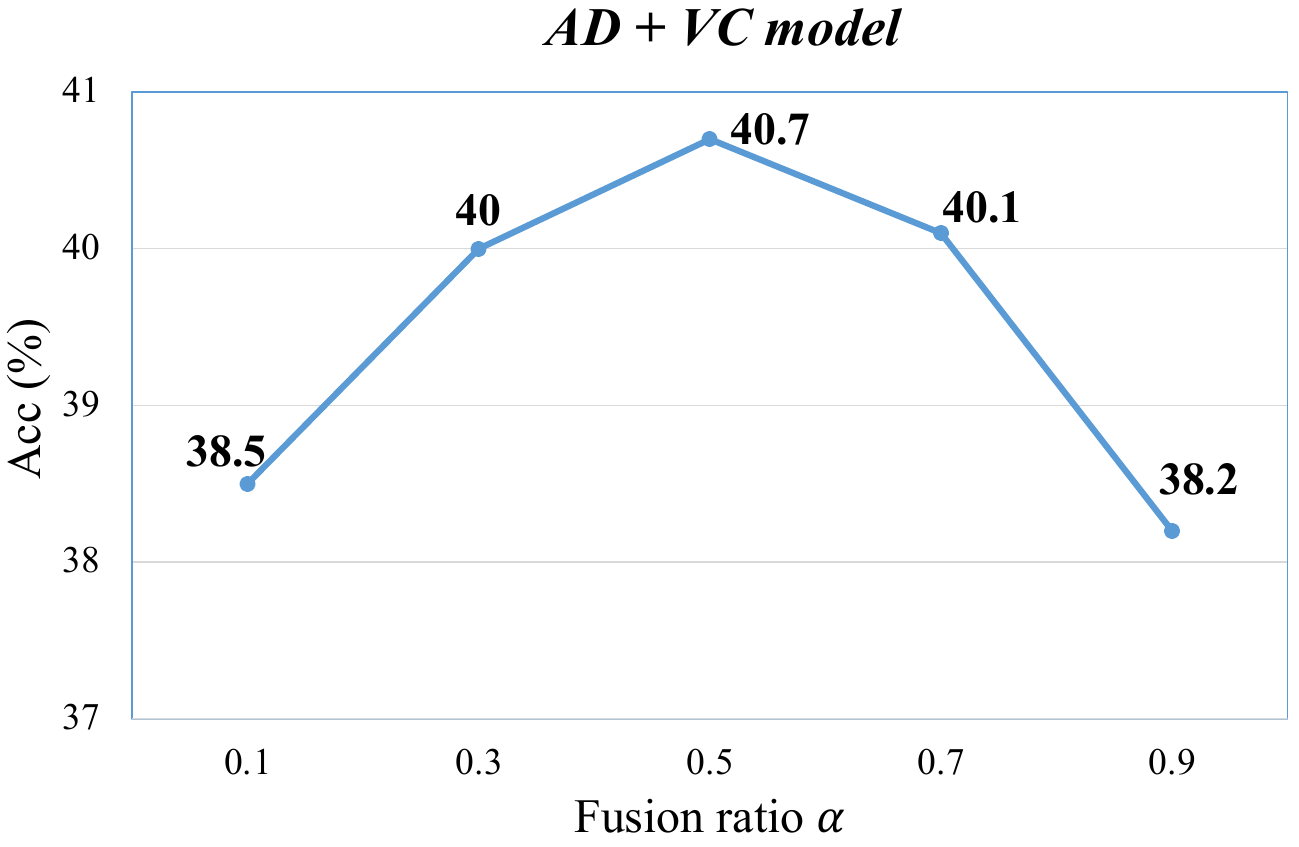}
\caption{The effects of different values of the fusion ratio $\alpha$ in Eq.~\ref{ED.SS.fusion1} and Eq.~\ref{ED.SS.fusion2}. Our \textit{AD + VC} model achieves best on the Kinetics-ZSAR dataset when the fusion ratio $\alpha$ is $0.5$.}
\label{fig:ablation_on_ED_SS_fusion_alpha}
\end{figure}

\noindent \textbf{- The fusion between the representation of action descriptions and video descriptions from ActionHub.} Experimental results in Table~\ref{tab:ablation_on_class_description} show that the action descriptions in our proposed ActionHub and existing action definitions are complementary in action semantics. To this end, we fuse the feature representation of video descriptions and action definitions, aiming to enrich the semantics of the class description feature for each action (see Eq.~\ref{ED.SS.fusion1} and Eq.~\ref{ED.SS.fusion2}). In Figure~\ref{fig:ablation_on_ED_SS_fusion_alpha}, we show the effects of different values of the fusion ratio $\alpha$ in Eq.~\ref{ED.SS.fusion1} and Eq.~\ref{ED.SS.fusion2}, which controls the fusion between features of video descriptions and action definitions. Our \textit{AD + VC} model achieves best when the fusion ratio $\alpha$ is $0.5$, which indicates the importance of both video descriptions and action definitions for representing the semantics of actions, and the complementarity of video descriptions and action definitions can be further demonstrated.

\vspace{0.1cm}

\begin{figure}[!h]
\centering
\includegraphics[width=0.8\columnwidth]{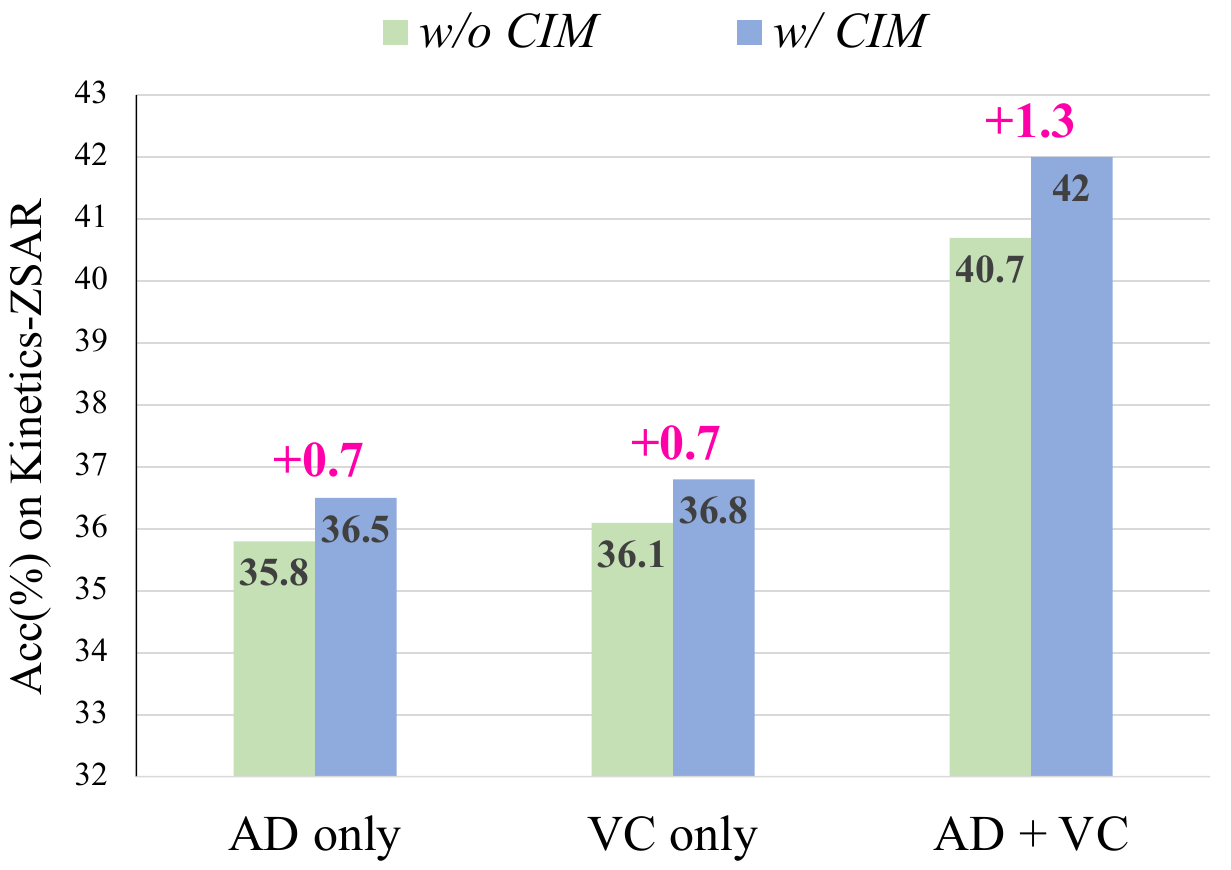}
\caption{The effectiveness of the Cross-action Invariance Mining (CIM) module. for all three types of class descriptions, \ie, \textit{AD-only} (only use action definitions), \textit{VC-only} (only use video descriptions), and \textit{AD + VC} (use action definitions and video descriptions), our CIM module can effectively improve the ZSAR performance of the model.}
\label{fig:ablation_on_using_CIM}
\end{figure}

\noindent \textbf{- The effectiveness of the Cross-action Invariance Mining (CIM) module.} This work proposes the CIM module to mine the cross-action invariant semantics for each action, which facilitates our model to learn a transferable alignment between video and text data of actions.
As shown in Figure~\ref{fig:ablation_on_using_CIM}, for all three types of class descriptions, \ie, \textit{AD-only} (only use action definitions), \textit{VC-only} (only use video descriptions), and \textit{AD + VC} (use action definitions and video descriptions), our CIM module can effectively improve the ZSAR performance of the model. And when both the action definitions and video descriptions (\ie, \textit{AD + VC}) are used as the class descriptions, the improvement brought by the CIM module is further boosted, compared with that of using action definitions or video descriptions alone as class descriptions. The above results demonstrate that based on the class descriptions with rich semantics, the proposed CIM module can effectively model the cross-action invariant semantics for recognizing unseen actions.

\subsection{Qualitative Results}

\begin{figure}[!t]
\centering
\includegraphics[width=1.0\columnwidth]{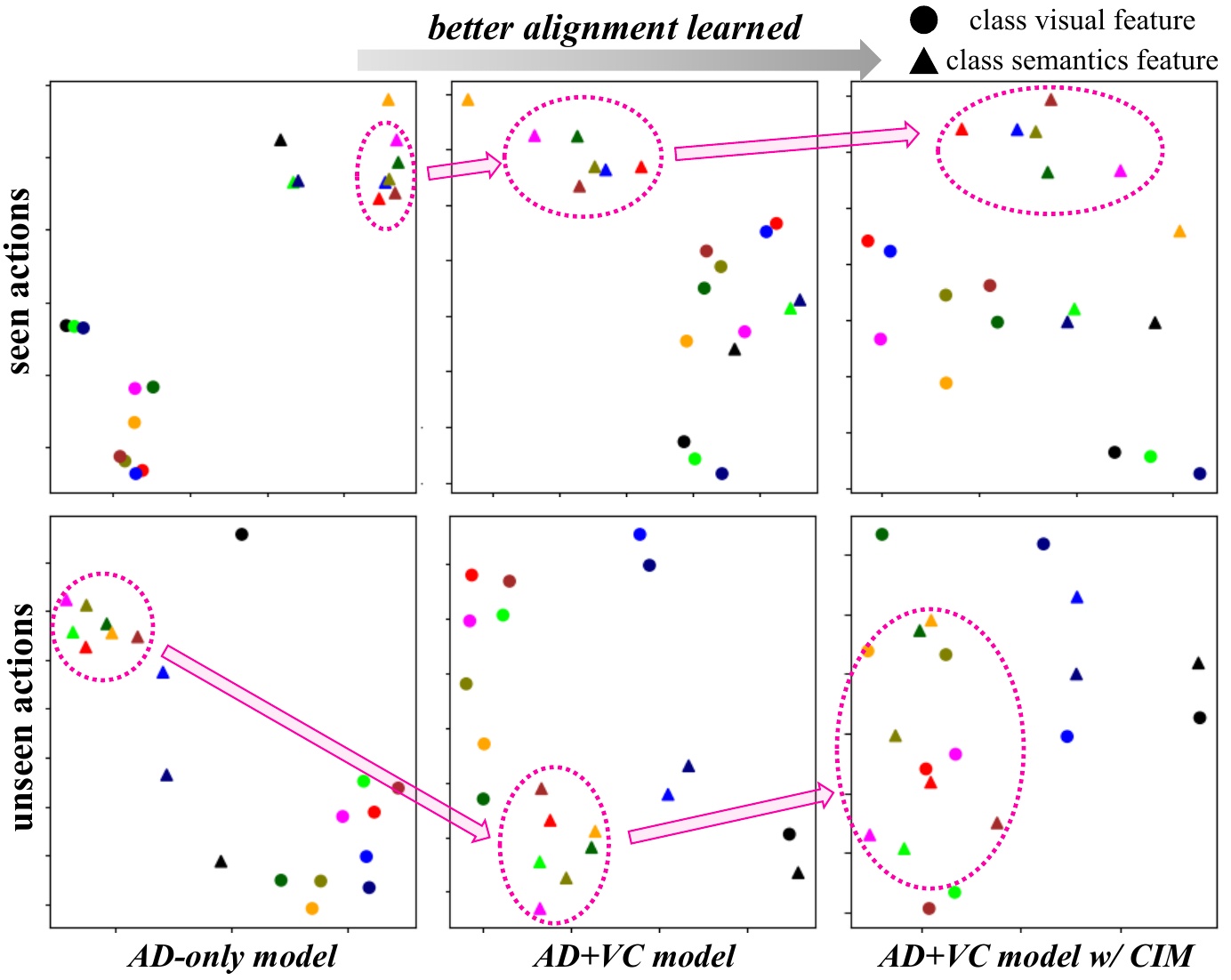}
\caption{The t-SNE visualizations of class visual/semantic features of 10 seen/unseen actions that are randomly selected from the Kinetics dataset. Each color represents an action class. The \textit{circle} $\bullet$ and \textit{triangle} $\blacktriangle$ represent the class visual features and class semantic features of the actions, respectively.}
\label{vis}
\end{figure}

To analyze how our method learns better representations, in Figure~\ref{vis}, we visualize the class visual/semantic features of $10$ seen/unseen actions that are randomly selected from the Kinetics-ZSAR dataset using t-SNE.
Each color in the figure represents an action class. The \textit{circle} $\bullet$ and \textit{triangle} $\blacktriangle$ represent the class visual feature and the class semantic feature of the action, respectively. 
The class visual feature of each action is obtained by averaging all video features of the action. The goal of the ZSAR models is to learn an alignment between the class visual feature and class semantic feature for each action. From Figure~\ref{vis} we can see that when only action definitions are used (\ie, the \textit{AD-only model}), there is severe overlap among the two types of features of different classes (see the pink-dashed 
circle in the left column of Figure~\ref{vis}), which leads to a misalignment between the visual and semantic features of the same action. 
When we use both action definitions and video descriptions to enrich the diversity of class descriptions (\ie, the \textit{AD+VC model}), the feature overlap is alleviated (see the larger pink-dashed circle of the same action classes in the second column of Figure~\ref{vis}). 
When the CIM module is further utilized (\ie, the \textit{AD+VC model w/ CIM}), we can observe a more uniform distribution of features as shown in the largest pink-dashed circle in the third column of Figure~\ref{vis}, which could lead to better feature alignment (\ie, the reduction of overlap between different action classes in the feature space will enhance the discrimination of each class, thus enabling better learning of feature alignment).
The above observations demonstrate the effectiveness of each component in our CoCo framework.

\begin{figure*}[!h]
\centering
\includegraphics[width=0.7\textwidth]{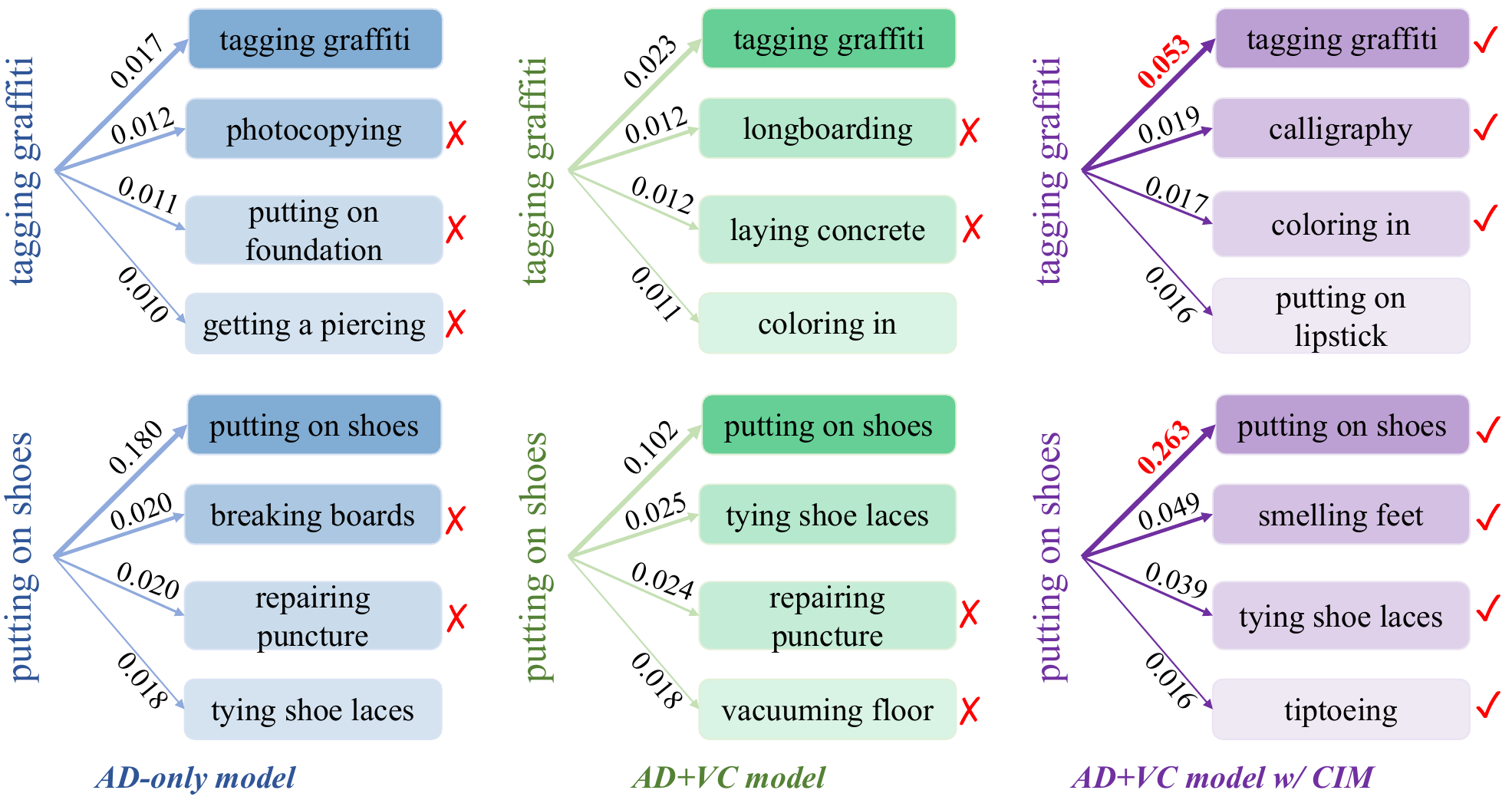}
\caption{The visualizations of the top-relevant actions in backward-pass of the Cross-action Invariance Mining module, which demonstrate the class semantic correlation in the Kinetics dataset. The \textit{AD+VC model w/ CIM} can associate actions with cross-action invariant semantics (see semantic-related actions marked with \textcolor{red}{\checkmark}), while \textit{AD-only model} and \textit{AD+VC model} fail to achieve this without CIM module (see semantic-unrelated actions marked with \textcolor{red}{\ding{55}}).}
\label{action_corr}
\end{figure*}

\begin{figure*}[!th]
\centering
\includegraphics[width=0.85\textwidth]{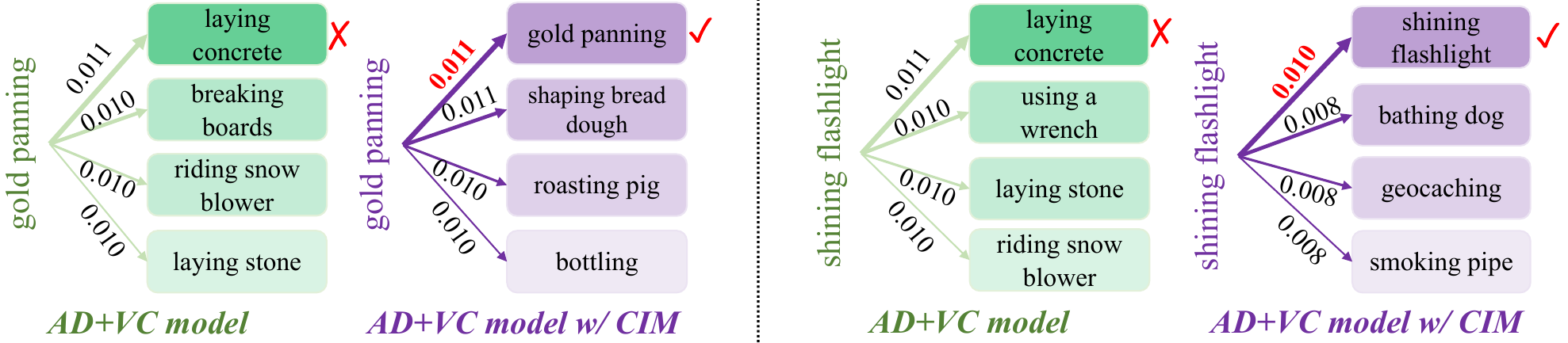}
\caption{Two action classes that have the least performance improvement after using the CIM module. The semantics-relevant classes are difficult to find, despite the use of CIM module. It is because these two actions are rare in daily life videos. However, we can find that the CIM module still helps our model maintain the original semantics of each action.}
\label{fig:R_bad_action_corr}
\end{figure*}

In Figure~\ref{action_corr}, we randomly select two actions from the Kinetics-ZSAR test set and show their top-$5$ correlated actions according to the attention map in backward-pass (Eq.~\ref{backward}) during action semantics reconstruction. Compared with the \textit{AD-only model} and the \textit{AD+VC model} that do not use the Cross-action Invariance Mining (CIM) module, our \textit{AD+VC model w/ CIM} using the constraint of class semantic cycle-consistency loss in the CIM module shows its superiority as follows. (i) For each action class, the correlation between its forward feature and original class semantic feature (see the red-highlighted attention value in Figure~\ref{action_corr}) is the highest in \textit{AD+VC model w/ CIM}, which means that the action classes can better maintain their semantic consistency in cycle-reconstruction process with our CIM module; (ii) The \textit{AD+VC model w/ CIM} can associate actions with cross-action invariant semantics (see semantic-related actions marked with \textcolor{red}{\checkmark} in Figure~\ref{action_corr}), while \textit{AD-only model} and \textit{AD+VC model} fail to achieve this without CIM module (see semantic-unrelated actions marked with \textcolor{red}{\ding{55}} in Figure~\ref{action_corr}), this phenomenon proves that the class semantics reconstruction in CIM module can make our model focus invariant semantics between actions. 

In addition, to investigate whether the proposed CIM module will introduce negative impacts, we select action classes that have the least performance improvement when using the CIM module. Following the same practice as Figure~\ref{action_corr}, we present two cases in Figure~\ref{fig:R_bad_action_corr}, in which we find that the semantics-relevant classes are difficult to find, despite the use of the CIM module. The reason is that these two actions are rare in daily life videos, which is one of the unsolved problems in ZSAR. However, we can find that the CIM module still helps our model maintain the original semantics of each action (see \textcolor{red}{\checkmark} in Figure \ref{fig:R_bad_action_corr}), while those without the CIM module cannot achieve this (see \textcolor{red}{\ding{55}} in Figure \ref{fig:R_bad_action_corr}).

\vspace{-0.35cm}

\section{Conclusion}

In this work, we propose the ActionHub dataset, which is the largest video action dataset with video descriptions to date, aiming to alleviate the cross-modality diversity gap between video and text data in zero-shot action recognition (ZSAR). To exploit the rich textual semantics of the proposed ActionHub dataset, we propose a Cross-modality and Cross-action Modeling (CoCo) framework, which includes a Dual Cross-modality Alignment module and a Cross-action Invariance Mining module. The Dual Cross-modality Alignment module utilizes both the action definitions and video descriptions from ActionHub to obtain class semantic features with rich semantics for feature alignment. To model the cross-action invariant representation, the Cross-action Invariance Mining module leverages a cycle-reconstruction process between class semantic features of different actions, expecting the reconstructed class semantic features of actions to maintain the corresponding semantics. Experimental results on exiting ZSAR benchmarks under two different learning protocols demonstrate the effectiveness of the proposed ActionHub dataset and CoCo framework.

% % use section* for acknowledgment
% \ifCLASSOPTIONcompsoc
%   % The Computer Society usually uses the plural form
%   \section*{Acknowledgments}
% \else
%   % regular IEEE prefers the singular form
%   \section*{Acknowledgment}
% \fi

% The authors would like to thank...

% % Can use something like this to put references on a page
% % by themselves when using endfloat and the captionsoff option.
% \ifCLASSOPTIONcaptionsoff
%   \newpage
% \fi

%%%%%%%%% REFERENCES
\bibliographystyle{IEEEtran}
\bibliography{egbib}

\begin{IEEEbiography}[{\includegraphics[width=1in,height=1.25in,clip,keepaspectratio]{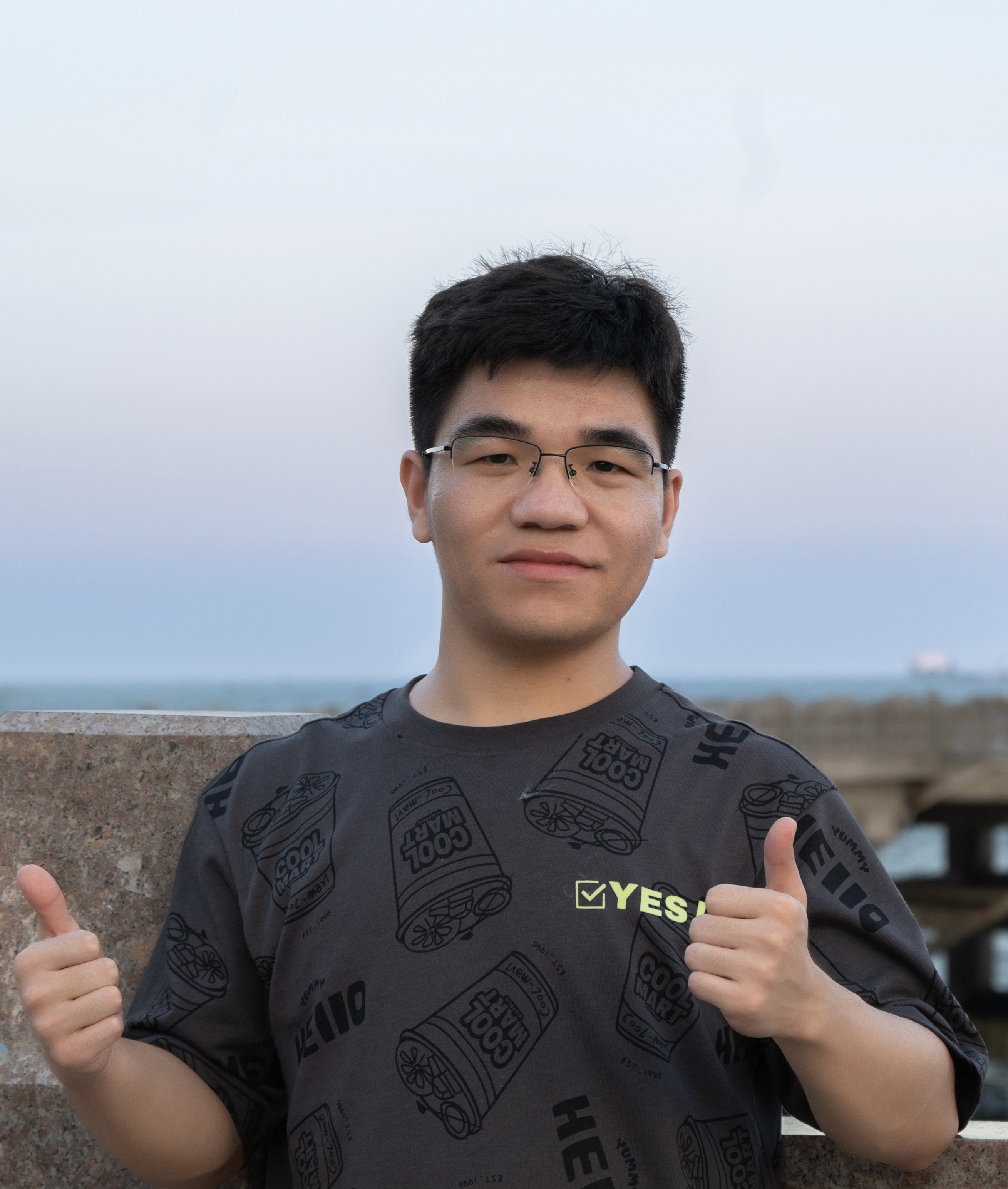}}]
{Jiaming Zhou} is a PhD student at Hong Kong University of Science and Technology, Guangzhou. He obtained his Bachelor's degree in Computer Science and Engineering, Sichuan University in 2020. Then he received his Master degree from Computer Science and Engineering, Sun Yat-Sen University in 2023. His research interests include video understanding and robot learning.
\end{IEEEbiography}

 \begin{IEEEbiography}[{\includegraphics[width=1in,height=1.25in,clip,keepaspectratio]{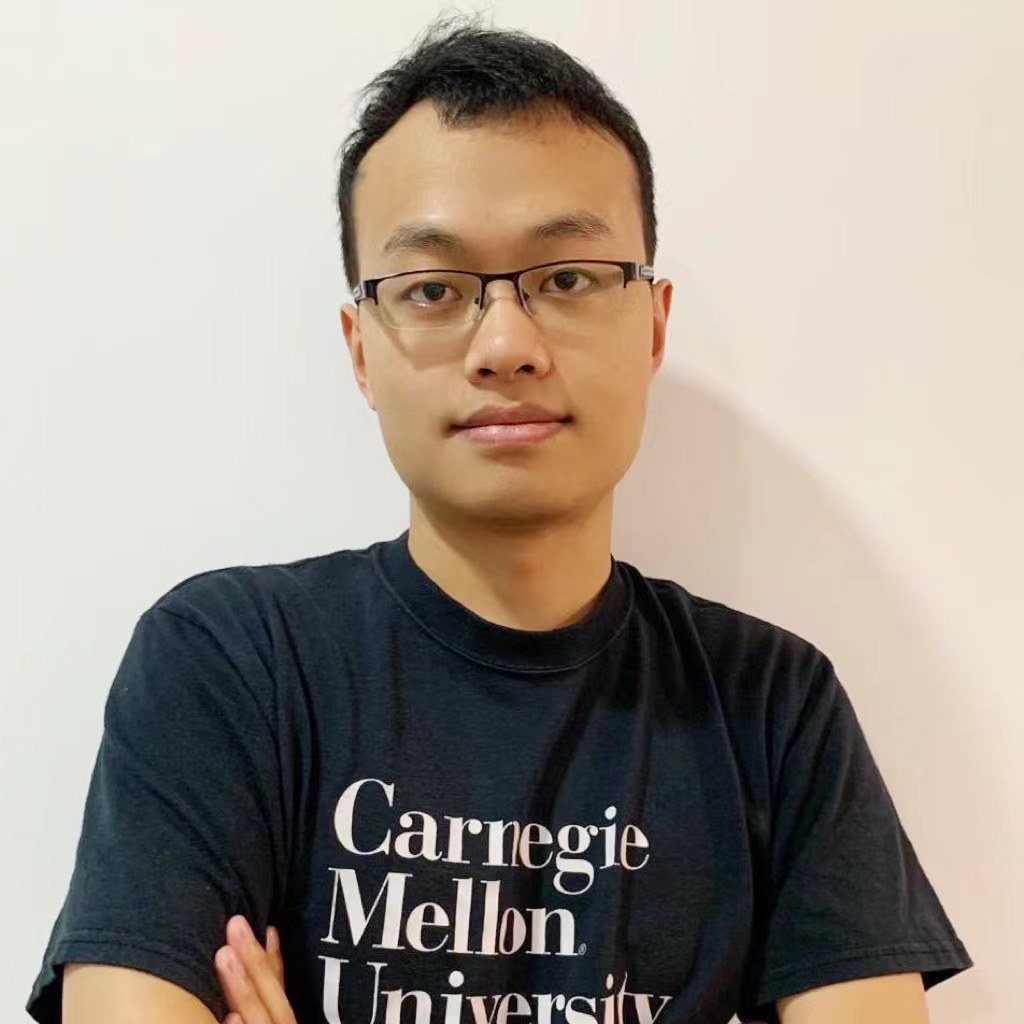}}]
 {Dr. Junwei Liang} is a tenure-track Assistant Professor (TTAP) at The Hong Kong University of Science and Technology (Guangzhou). He is also affiliated with HKUST CSE. He was a senior researcher at Tencent Youtu Lab. Before that, he received his Ph.D. from Carnegie Mellon University. He received the Baidu Scholarship, Yahoo Fellowship and ICCV Doctoral Consortium Award. He received the Rising Star Award at the World AI Conference in 2020. He is the winner of several public safety video analysis competitions, including NIST ASAPS and TRECVID ActEV. His work has helped and been reported by major news agencies like the Washington Post and New York Times. His research interests include human trajectory forecasting, action recognition, and large-scale computer vision.
 \end{IEEEbiography}

\begin{IEEEbiography}[{\includegraphics[width=1in,height=1.25in,clip,keepaspectratio]{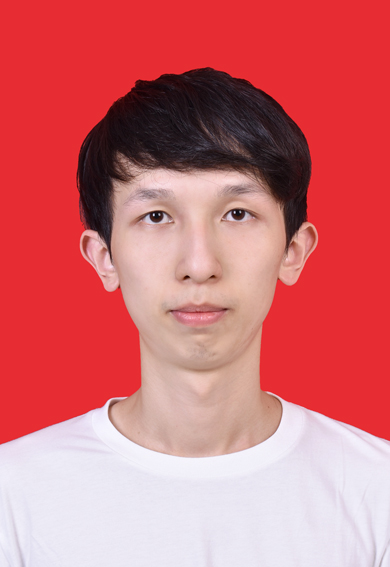}}]
{Kun-Yu Lin} received a B.S. and M.S. degree from the School of Data and Computer Science, Sun Yat-sen University, China. He is currently a PhD student in the School of Computer Science and Engineering, Sun Yat-sen University. His research interests include computer vision and machine learning.
\end{IEEEbiography}

\begin{IEEEbiography}[{\includegraphics[width=1in,height=1.25in,clip,keepaspectratio]{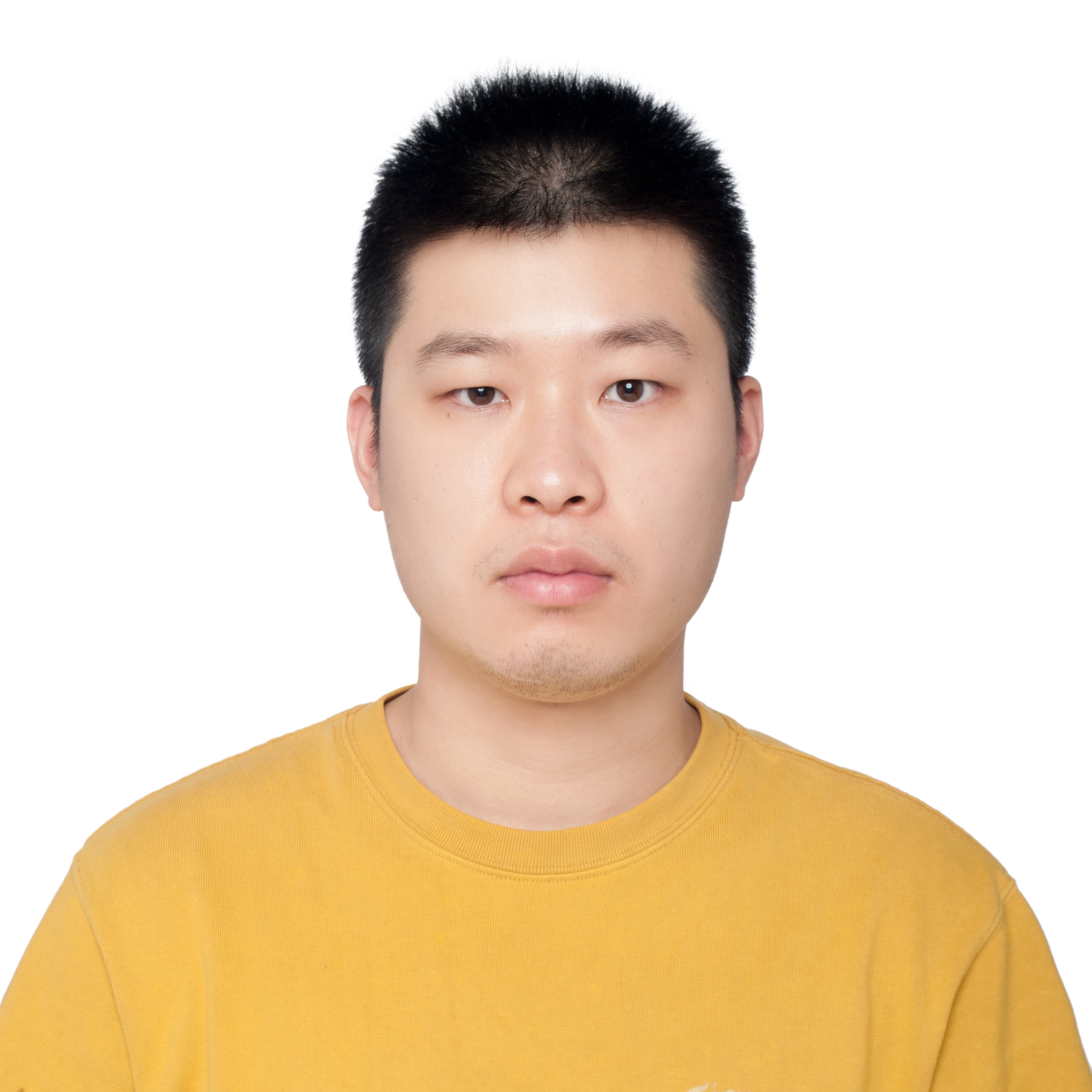}}]
{Jinrui Yang} received his B.E. degree in Software Engineering from Sichuan University, Chengdu, China, in 2019, and his M.E. degree from the School of Computer Science and Engineering at Sun Yat-sen University, Guangzhou, China, in 2021. He also worked at Tencent Youtu Lab as a researcher from 2021 to 2023. Currently, he is a Ph.D. student in the Computer Science and Engineering Department at the University of California, Santa Cruz, beginning in 2023. His research interests include multimodal learning and large language models.
\end{IEEEbiography}

\begin{IEEEbiography}[{\includegraphics[width=1in,height=1.25in,clip,keepaspectratio]{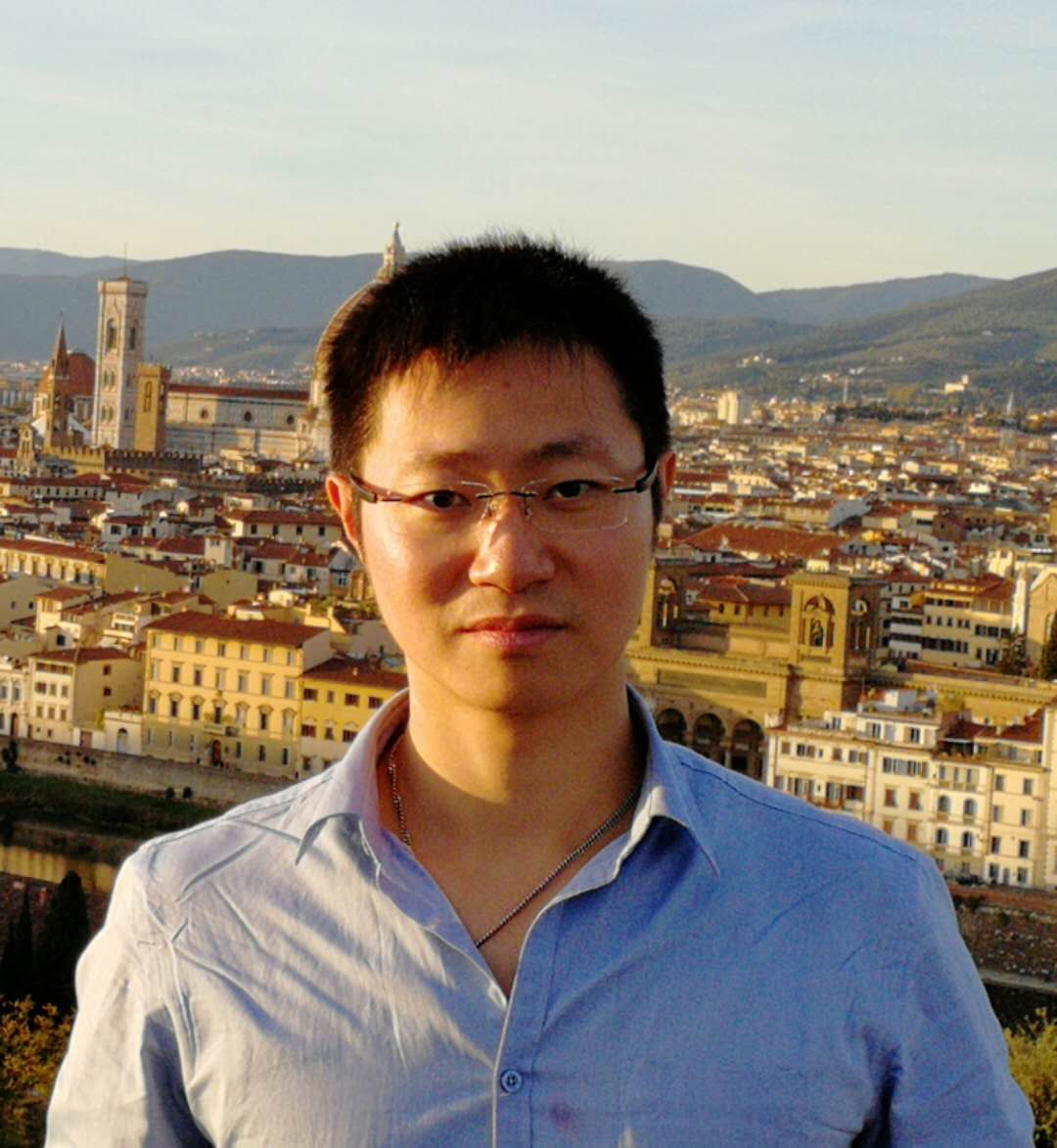}}]
{Dr. Wei-Shi Zheng} is now a Professor with Sun Yat-sen University. His research interests include person/object association and activity understanding in visual surveillance, and the related large-scale machine learning algorithm. He has ever served as area chairs of ICCV, CVPR, ECCV, NeurIPS, BMVC, IJCAI and AAAI. He is an associate editor of IEEE-TPAMI and Pattern Recognition. He has ever joined Microsoft Research Asia Young Faculty Visiting Programme. He is a Cheung Kong Scholar Distinguished Professor, a recipient of the Excellent Young Scientists Fund of the National Natural Science Foundation of China, and a recipient of the Royal Society-Newton Advanced Fellowship of the United Kingdom.
\end{IEEEbiography}

% that's all folks
\end{document}